\documentclass[letterpaper, 10 pt, conference]{ieeeconf}  

\IEEEoverridecommandlockouts                              

\usepackage[linesnumbered,ruled,vlined]{algorithm2e}
\usepackage{algorithmic}
\usepackage{graphics} 
\graphicspath{{figures/}}
\usepackage{epsfig} 
\usepackage{times} 
\usepackage{siunitx}
\usepackage{amsmath} 
\usepackage{amssymb}  
\usepackage{hyperref}
\usepackage[capitalise]{cleveref}
\Crefname{algorithm}{Alg.}{Algs.}
\Crefname{section}{Sec.}{Secs.}
\Crefname{equation}{Eq.}{Eqs.}
\usepackage[colorinlistoftodos]{todonotes}
\usepackage{booktabs}
\usepackage[
    sorting=none,
    giveninits=true,
    maxbibnames=9,
    maxcitenames=2,backend=biber
    ]{biblatex}
\title{\LARGE \bf
CLiFF-LHMP: Using Spatial Dynamics Patterns for\break Long-Term Human Motion Prediction
}

\author{
Yufei Zhu$^{1}$, Andrey Rudenko$^{2}$, Tomasz P. Kucner$^{3}$, \\ Luigi Palmieri$^{2}$, Kai O. Arras$^{2}$, Achim J. Lilienthal$^{1,4}$, Martin Magnusson$^{1}$%
\thanks{$^{1}$AASS MRO lab, Örebro University, Sweden {\tt\small yufei.zhu@oru.se}}%
\thanks{$^{2}$Bosch Corporate Research, Robert Bosch GmbH, Stuttgart, Germany {\tt\small andrey.rudenko@bosch.com}}%
\thanks{$^{3}$Finnish Center for Artificial Intelligence, School of Electrical Engineering, Aalto University, Finland}%
\thanks{$^{4}$TU Munich, Germany}%
\thanks{This work has received funding from the European Union’s Horizon 2020 research and innovation programme under grant agreement No 101017274 (DARKO).}
}

\begin{document}

\maketitle
\thispagestyle{empty}
\pagestyle{empty}

\begin{abstract}

Human motion prediction is important for mobile service robots and intelligent vehicles to operate safely and smoothly around people. The more accurate predictions are, particularly over extended periods of time, the better a system can, e.g., assess collision risks and plan ahead. In this paper, we propose to exploit \emph{maps of dynamics} (MoDs, a class of general representations of place-dependent spatial motion patterns, learned from prior observations) for long-term human motion prediction (LHMP). We present a new MoD-informed human motion prediction approach, named CLiFF-LHMP, which is data efficient, explainable, and insensitive to errors from an upstream tracking system. Our approach uses CLiFF-map, a specific MoD trained with human motion data recorded in the same environment. We bias a constant velocity prediction  with samples from the CLiFF-map to generate multi-modal trajectory predictions. In two public datasets we show that this algorithm outperforms the state of the art for predictions over very extended periods of time, achieving 45\% more accurate prediction performance at 50s compared to the baseline.

\end{abstract}


\section{INTRODUCTION}
Accounting for long-term human motion prediction (LHMP) is an important task for autonomous robots and vehicles to operate safely in populated environments \cite{rudenko2020human}. Accurate prediction of future trajectories of surrounding people over longer periods of time is a key skill to improve motion planning, tracking, automated driving, human-robot interaction, and surveillance. Long-term predictions are useful to associate observed tracklets in sparse camera networks, or inform the robot of the long-term environment dynamics on the path to its goal \cite{palmieri2017kinodynamic, swaminathan2018down}, for instance when following a group of people. Very long-term predictions are useful for global motion planning to produce socially-aware unobtrusive trajectories, and for coordinating connected multi-robot systems with sparse perception fields. 

Human motion is complex and may be influenced by several hard-to-model factors, including social rules and norms, personal preferences, and subtle cues in the environment that are not represented in geometric maps. Accordingly, accurate motion prediction is very challenging \cite{rudenko2020human}.
Prediction on the very long-term scale (i.e., over \SI{20}{\second} into the future) is particularly hard as complex, large-scale environments influence human motion in a way that cannot be summarized and contained in the current state of the moving person or the observed interactions but rather have to be modelled explicitly~\cite{Rudenko2018iros}. 


In this paper, we examine and address the novel task of very long-term human motion prediction \cite{tanke2021intention}, aiming to predict human trajectories for up to \SI{50}{\second} into the future.
Prior works have addressed human motion prediction using physics-, planning- and pattern-based approaches \cite{rudenko2020human}. The majority of existing approaches, however, focuses on relatively short prediction horizons (up to \SI{10}{\second}) \cite{rudenko2022atlas} and the popular ETH-UCY benchmark uses \SI{4.8}{\second} \cite{rudenko2020human,amirian2019social,pang2021trajectory,gu2022stochastic}. 

\begin{figure}
\centering
\includegraphics[width=\linewidth]{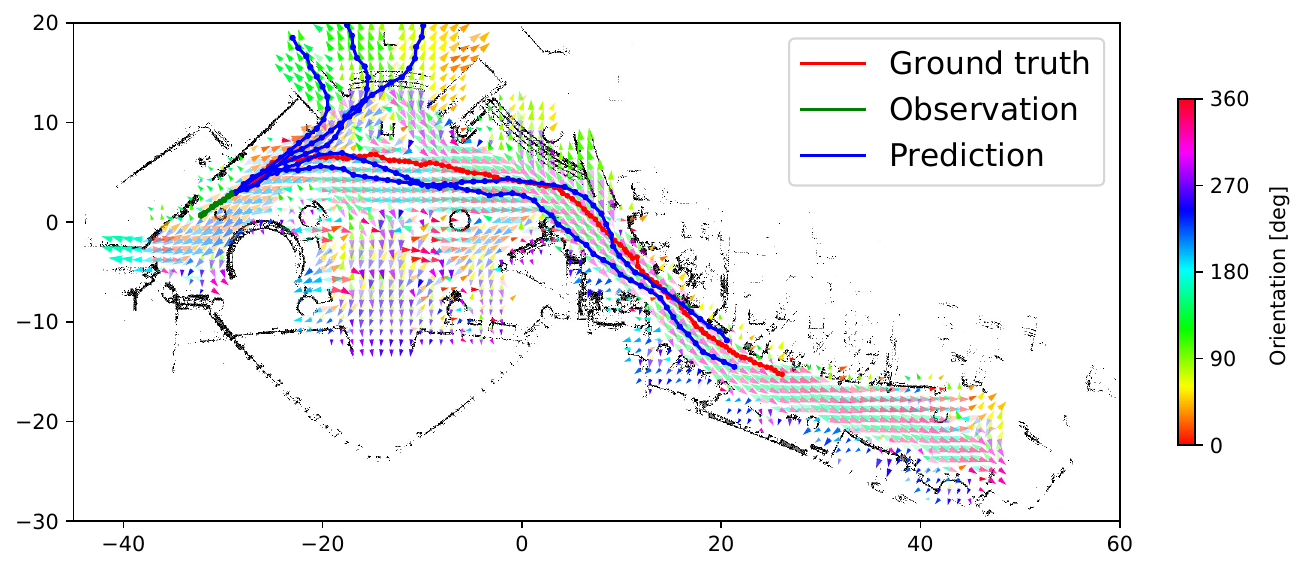}
\caption{Long-term (\SI{50}{\second}) motion prediction result obtained with CLiFF-LHMP for one person in the ATC dataset. \textbf{Red} line: ground truth trajectory. \textbf{Green} line: observed trajectory. \textbf{Blue} lines: predicted trajectories. The CLiFF-map is shown with colored arrows.} 
\label{fig:predict-example-main}
\vspace*{-5mm}
\end{figure}

To predict very long-term human motion, we exploit \emph{maps of dynamics} (MoDs) that encode human dynamics as a feature of the environment. There are several MoD approaches for mapping velocities \cite{molina2018modelling, krajnik2017fremen, kucner2017enabling, zhi2019spatiotemporal, kucner2020probabilistic}. In this work, we use Circular Linear Flow Field map (CLiFF-map) \cite{kucner2017enabling}, which captures multimodal statistical information about human flow patterns in a continuous probabilistic representation over velocities. 
The motion patterns represented in a CLiFF-map implicitly avoid collisions with static obstacles and follow the topological structure of the environment, e.g., capturing the dynamic flow through a hall into a corridor (see \cref{fig:predict-example-main}). 
In this paper we present a novel, MoD-informed prediction approach (CLiFF-LHMP)\footnote{The approach is available at \url{https://github.com/test-bai-cpu/CLiFF-LHMP}} that predicts stochastic trajectories by sampling from a CLiFF-map to guide a velocity filtering model \cite{rudenko2022atlas}. Examples of prediction results are shown in \cref{fig:predict-example-main}.

In qualitative and quantitative experiments we demonstrate our CLiFF-LHMP approach is 45\% more accurate than the baseline at \SI{50}{\second}, with average displacement error (ADE) below \SI{5}{\metre} up to \SI{50}{\second}. In contrast to prior art in long-term environment-aware motion prediction \cite{Rudenko2018iros}, our method does not make any assumptions on the optimality of human motion and instead generalizes the features of human-space interactions from the learned MoD. Furthermore, our method does not require a list of goals in the environment as input, in contrast to prior planning-based prediction methods. Finally, our method can flexibly estimate the variable time end-points of human motion, predicting both short- and long-term trajectories, in contrast to the prior art which always predicts up to a fixed prediction horizon.


The paper is structured as follows: we review related work in \cref{section-relatedwork}, describe the proposed approach in \cref{section-method}, present our evaluation in \cref{section-experiments}, discuss the results in \cref{section-results} and conclude in \cref{section-conclusions}.






\section{RELATED WORK} \label{section-relatedwork}

Human motion prediction has been studied extensively in recent years. With different prediction horizons, the human motion prediction problem can be divided into short-term (1--\SI{2}{\second}), long-term (up to \SI{20}{\second}) \cite{rudenko2020human}, and very long-term (which we define as over \SI{20}{\second}). Several approaches address long-term motion prediction, e.g., full-body motion \cite{tanke2021intention} or in the context of vehicle routing and GPS positioning \cite{cheng2010geodtn,xiao2018goi}, but, to the best of our knowledge, very long-term prediction of dense navigation trajectories has not been addressed before.

One approach to predict long-term human motion is to account for various semantic attributes of the static environment. For instance, prior knowledge of potential goals in the environment can be used in planning-based methods. \textcite{ziebart2009planning,karasev2016intent} propose planning MDP-based approaches for long-term goal-directed global motion prediction. \textcite{Rudenko2018iros} extends this line of work by accounting for local social interactions, which is shown to outperform prior art in the long-term map-aware perspective. 

Another popular approach to make long-term predictions is using clustering to represent observed long-term motion patterns, e.g., using expectation-maximization~\cite{bennewitz2005learning}. \textcite{chen2008pedestrian} use constrained gravitational clustering 
for dynamically grouping the observed trajectories, learning also how motion patterns change over time. \textcite{bera2016glmp} learn global and local motion patterns using Bayesian inference in real-time. One shortcoming of clustering-based methods is that they depend on complete trajectories as input. In many cases, e.g. in cluttered environments or from a first-person perspective \cite{dondrup2015tracker}, it is difficult to observe long trajectories, or cluster shorter tracklets and incomplete trajectories in a meaningful way.

Clustering-based methods directly model the distribution over full trajectories and are non-sequential. By contrast, transition-based approaches \cite{thompson2009probabilistic, wang2015modeling, ballan2016knowledge, kucner2013conditional, Jari2012iMac} describe human motion with causally conditional models 
and generate sequential predictions from learned local motion patterns.


Further, there are physics-based approaches that build a kinematic model without considering other forces that govern the motion. The constant velocity model (CVM) is a simple yet potent approach to predict human motion. \textcite{scholler2019simpler} have shown CVM to outperform several state-of-the-art neural predictors at the \SI{4.8}{\second} prediction horizon. On the other hand, CVM is not reliable for long-term prediction as it ignores all environment information.

Finally, many neural network approaches for motion prediction have been presented in recent years, based on LSTMs \cite{alahi2016social}, GANs \cite{sadeghian2018sophie}, CNNs \cite{mohamed2020social}, CVAEs \cite{salzmann2020trajectronpp} and transformers \cite{giuliari2021transformer}. Most of these approaches focus on learning to predict stochastic interactions between diverse moving agents in the short-term perspective in scenarios where the effect of the environment topology and semantics is minimal. Our approach, on the other hand, targets specifically the long-term perspective, where the environment effects become critical for making accurate predictions.

Our approach to motion prediction leverages maps of dynamics (MoDs), which encode motion as a feature of the environment by building spatio-temporal models of the patterns followed by dynamic objects (such as humans) in the environment \cite{kucner2020probabilistic, kucner2017enabling}. There are several approaches for building maps of dynamics from observed motion. Some MoDs represent human dynamics in occupancy grid maps~\cite{wang2015modeling}. Another type of MoDs clusters human trajectories as mentioned above \cite{bennewitz2005learning}. \textcite{chen2016augmented} present an approach that uses a dictionary learning algorithm to develop a part-based trajectory representation. 

The above mentioned MoDs encode the direction but not the speed of motion. MoDs can also be based on mapping sparse velocity observations into flow models, which has the distinct advantage that the MoD can be built from incomplete or spatially sparse data.
An example of this class of MoDs is the probabilistic Circular-Linear Flow Field map (CLiFF-map)~\cite{kucner2017enabling} that we use in this paper. CLiFF-map uses a Gaussian mixture model (GMM) to describe multimodal flow patterns at each location. In this paper, we use sampled directions from the CLiFF-map to predict stochastic long-term human motion.


A method similar to ours is presented in~\textcite{BARATA2021107631}.
It constructs a vector field that represents the most common direction at each point and predicts human trajectories by inferring the most probable sequence through this vector field.
By contrast, our approach uses a probabilistic vector field that represents speed and direction jointly in a multimodal distribution. 
Further, the evaluation in~\textcite{BARATA2021107631} assumes a fixed prediction horizon of \SI{4.8}{\second}, 
whereas we show our approach to estimate human motion more accurately than the state of the art for up to \SI{50}{\second}.

\section{METHOD} \label{section-method}

In this section, we first describe the CLiFF-map representation for site-specific motion patterns (\cref{method-A}) and then present the CLiFF-LHMP approach for single-agent long-term motion prediction exploiting the information accumulated in a CLiFF-map (\cref{method-B}).

\subsection{Circular-Linear Flow Field Map (CLiFF-map)} \label{method-A}

\begin{figure*}[t]
\centering
\includegraphics[width=.22\linewidth]{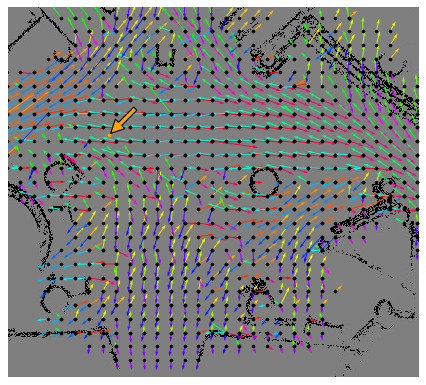}
\includegraphics[width=.22\linewidth]{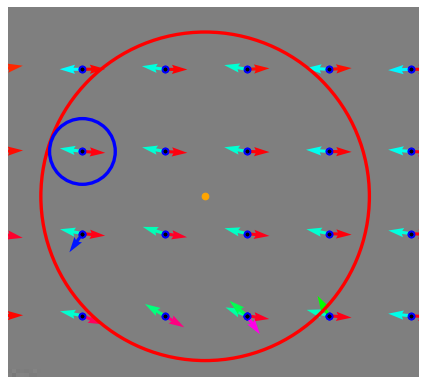}
\includegraphics[width=.25\linewidth]{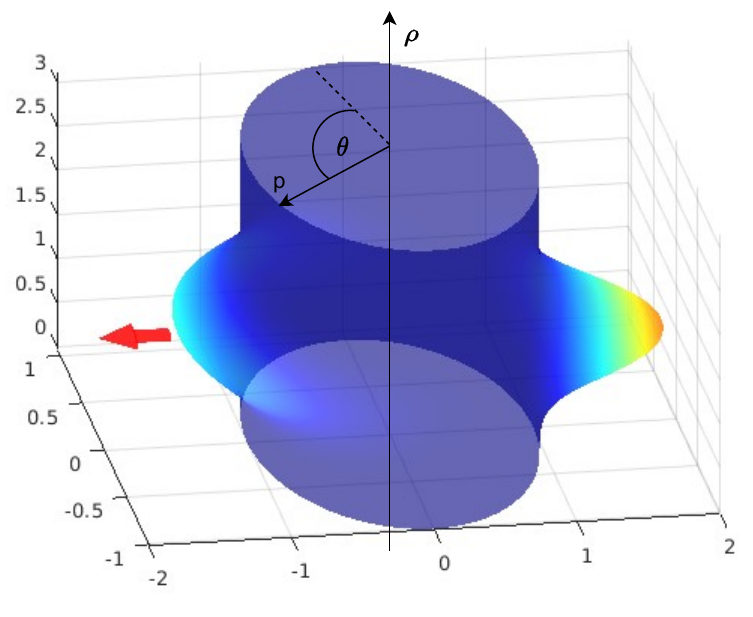}
\includegraphics[width=.28\linewidth]{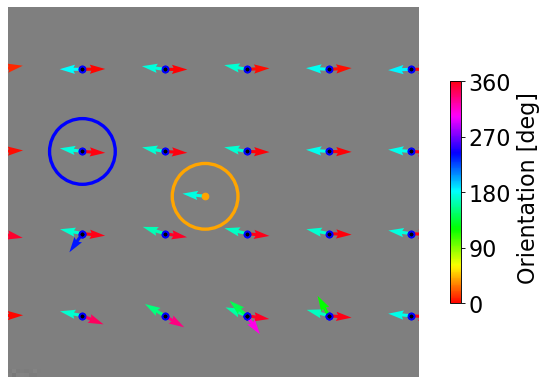} \\
\includegraphics[width=.80\linewidth]{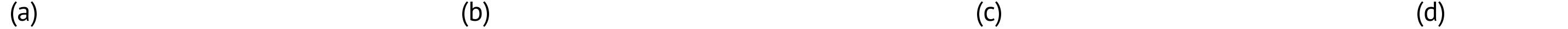}

\caption{Steps of sampling a direction $\theta_s$ from the CLiFF-map. {\bf (a)} CLiFF-map built from the ATC data. The location to sample from is marked with an orange arrow. {\bf (b)} Selection of SWGMMs in the CLiFF-map: The red circle contains all SWGMMs within $r_s$ distance to the sampling location. From these SWGMMs, the SWGMM with the highest motion ratio is selected (marked with a blue circle). {\bf (c)} The SWGMM distribution in the selected location wrapped on a unit cylinder. The speed is represented by the position along the $\rho$ axis and the direction is $\theta$. The probability is represented by the distance from the surface of the cylinder. A velocity vector (marked with a red arrow) is sampled from this SWGMM. {\bf (d)} The direction value $\theta_s$ of the sampled velocity is shown in the sampled direction and marked with an orange circle.} \label{fig:sample}
\vspace*{-6mm}
\end{figure*}

To predict human trajectories we exploit the information about local flow patterns represented in a CLiFF-map as a multimodal, continuous distribution over velocities.
CLiFF-map~\cite{kucner2017enabling} is a probabilistic framework for mapping velocity observations (independently of their underlying physical processes), i.e., essentially a generalization of a vector field into a Gaussian mixture field.
Each location in the map is associated with a Gaussian mixture model (GMM).
A CLiFF-map represents motion patterns based on local observations and estimates the likelihood of motion at a given query location. 

CLiFF-maps represent speed and direction jointly as velocity $\mathbf{V} = [\theta, \rho]^T$ using direction $\theta$ and speed $\rho$, where $\rho \in \mathbb{R}^+$, $\theta \in [0,2\pi)$.
%
As the direction $\theta$ is a circular variable and the speed is linear, a mixture of \emph{semi-wrapped} normal distributions (SWNDs) is used in CLiFF-map. At a given location, the semi-wrapped probability density function (PDF) over velocities can be visualized as a function on a cylinder. Direction values $\theta$ are wrapped on the unit circle and the speed $\rho$ runs along the length of the cylinder. An SWND $\mathcal{N}^{SW}_{\boldsymbol{\Sigma}, \boldsymbol{\mu}}$ is formally defined as 
$\mathcal{N}^{SW}_{\boldsymbol{\Sigma}, \boldsymbol{\mu}}(\mathbf{V}) = \sum_{k \in \mathbb{Z}} \mathcal{N}_{\boldsymbol{\Sigma}, \boldsymbol{\mu}}( [\theta,  \rho]^T + 2\pi [ k,  0 ]^T )$, 
where $\boldsymbol{\Sigma}, \boldsymbol{\mu}$ denote the covariance matrix and mean value of the directional velocity $(\theta, \rho)^T$, and $k$ is a winding number. Although $k \in \mathbb{Z}$, the PDF can be approximated adequately by taking $k \in \{-1, 0, 1\}$ for practical purposes \cite{jupp2018circlulalrbook}.
To preserve the multi-modal characteristic of the flow, a semi-wrapped Gaussian mixture model (SWGMM) is used, which is a PDF represented as a weighted sum of $J$ SWNDs: $p(\mathbf{V} | \mathbf{\xi}) = \sum_{j=1}^{J}\pi_{j}\mathcal{N}^{SW}_{\boldsymbol{\Sigma_j}, \boldsymbol{\mu_j}}(\mathbf{V})$, where $\boldsymbol{\xi} = \{ \xi_j = (\boldsymbol{\mu}_j, \boldsymbol{\Sigma}_j, \pi_j) | j \in \mathbb{Z}^+ \}$ denotes a finite set of components of the SWGMM, and  $\pi_j$ denotes the mixing factor and satisfies $0 \leq \pi_j \leq 1$.



\subsection{
Human Motion Prediction Using CLiFF-map} \label{method-B}
We frame the task of predicting a person's future trajectory as inferring a sequence of future states. The algorithm is presented in \cref{alg:LHMPAlgo}. With the input of an observation history of $O_p$ past states of a person and a CLiFF-map $\Xi$, the algorithm predicts $T_p$ future states. The length of the observation history is $O_s \in \mathbb{R}^+$~\SI{}{\second}, equivalent to $O_p > 0$ observation time steps.
%
%
%
%
With the current time-step denoted as the integer $t_0 \geq 0$, 
the sequence of observed states is $\mathcal{H} = \langle s_{t_{0} - 1},..., s_{t_{0} - O_p} \rangle$, where $s_t$ is the state of a person at time-step $t$. 
A state is represented by 2D Cartesian coordinates $(x,y)$, speed $\rho$ and direction $\theta$:
$s = (x,y,\rho, \theta)$.

From the observed sequence $\mathcal{H}$, we derive the observed speed $\rho_{\mathrm{obs}}$ and direction $\theta_{\mathrm{obs}}$ at time-step $t_0$ (line 2 of \cref{alg:LHMPAlgo}). Then the current state becomes $s_{t_0} = (x_{t_0},y_{t_0},\rho_{\mathrm{obs}},\theta_{\mathrm{obs}})$ (line 3 of \cref{alg:LHMPAlgo}). The values of $\rho_{\mathrm{obs}}$ and  $\theta_{\mathrm{obs}}$ are calculated as a weighted sum of the finite differences in the observed states, as in the recent ATLAS benchmark \cite{rudenko2022atlas}. With the same parameters as in~\cite{rudenko2022atlas}, the sequence of observed velocities is weighted with a zero-mean Gaussian kernel with $\sigma = 1.5$ to put more weight on more recent observations, such that $\rho_{\mathrm{obs}} = \sum_{t=1}^{O_p}v_{t_0 - t}g(t)$ and $\theta_{\mathrm{obs}} = \sum_{t=1}^{O_p}\theta_{t_0 - t}g(t)$, where $g(t) = 
(\sigma\sqrt{2\pi}e^{\frac{1}{2}(\frac{t}{\sigma})^2})^{-1}$.


Given the current state $s_{t_0}$, we estimate a sequence of future states.
Similar to past states, future states are predicted within a time horizon $T_s \in \mathbb{R}^+$~\SI{}{\second}. $T_s$ is equivalent to 
$T_p > 0$ 
prediction time steps, assuming a constant time interval $\Delta t$ between two predictions. Thus, the prediction horizon is $T_s = T_p \Delta t$. The predicted sequence is then denoted as $\mathcal{T} = \langle s_{t_0+1}, s_{t_0+2},...,s_{t_0+T_p} \rangle$.

\begin{algorithm}[t]
\small
    \KwIn{$\mathcal{H}$, $x_{t_0}$, $y_{t_0}, \Xi$}
    \KwOut{$\mathcal{T}$}
	$\mathcal{T} = \{\}$ \
	
	$\rho_\mathrm{obs}, \theta_\mathrm{obs} \leftarrow $  getObservedVelocity($\mathcal{H}$) \
	
	$s_{t_0} = (x_{t_0},y_{t_0},\rho_{\mathrm{obs}},\theta_{\mathrm{obs}})$ \
	
	\For { $t= t_{0}+1$, ..., $t_{0}+T_p$}{
	
	    $x_t, y_t \leftarrow $ getNewPosition($s_{t\textendash1}$) \
	    
	    
		$\theta_s$ $ \leftarrow $ sampleDirectionFromCLiFFmap($x_t, y_t, \Xi$)\
	    
		($\rho_t$, $\theta_t$) $ \leftarrow $ predictVelocity($\theta_{s}$, $\rho_{t\textendash1}$, $\theta_{t\textendash1}$)\
		
		$s_t \leftarrow (x_t, y_t, \rho_{t}, \theta_t)$
		
		$\mathcal{T} \leftarrow \mathcal{T} \cup s_t$ \
    }
    \Return $\mathcal{T}$ \

 \caption{CLiFF-LHMP}
\label{alg:LHMPAlgo}
\end{algorithm}

To estimate $\mathcal{T}$, for each prediction time step, we sample a direction from the CLiFF-map at the current position ($x_t$, $y_t$) to bias the prediction with the learned motion patterns represented by the CLiFF-map. The main steps for each iteration are shown in lines 5--9 of \cref{alg:LHMPAlgo}.

For each iteration, we first compute the predicted position $(x_t, y_t)$ at time step $t$ from the state at the previous time step (line 5 of \cref{alg:LHMPAlgo}):
\begin{equation}
\begin{gathered}
    x_{t} = x_{t-1} + \rho_{t-1}\cos{\theta_{t-1}} \Delta t,\\
    y_{t} = y_{t-1} + \rho_{t-1}\sin{\theta_{t-1}} \Delta t,\\
\end{gathered}
\end{equation}

Afterwards, we estimate the new speed and direction using constant velocity prediction biased by the CLiFF-map. The bias impacts only the estimated direction of motion, speed is assumed to be unchanging.

To estimate direction at time $t$, we sample a direction from the CLiFF-map at location $(x_t, y_t)$ in the function \texttt{sampleDirectionFromCLiFFmap()} (line 6 of \cref{alg:LHMPAlgo}). \Cref{alg:sample} outlines its implementation.
%
\begin{algorithm}[t]
\small
    \KwIn{$x$, $y$, $\Xi$}
    \KwOut{$\theta_s$}
	$\Xi_{\mathrm{near}} \leftarrow $ getNearSWGMMs($x, y, \Xi$) \
	
	$\xi \leftarrow $ selectSWGMM($\Xi_{\mathrm{near}}$) \
	
	$\theta_s$ $\leftarrow $ sampleDirectionFromSWGMM($\xi$) \

    \Return $\theta_s$
 \caption{sampleDirectionFromCLiFFmap($x, y, \Xi$)}
\label{alg:sample}
\end{algorithm}
The inputs of \cref{alg:sample} are: the sample location $(x, y)$ and the CLiFF-map $\Xi$ of the environment. The sampling process is illustrated in \cref{fig:sample}. 
To sample a direction at location $(x,y)$, 
from $\Xi$, we first get the SWGMMs $\Xi_{\mathrm{near}}$ whose distances to $(x,y)$ are less than the sampling radius $r_s$ (line 1 of \cref{alg:sample}). In a CLiFF-map, each SWGMM is associated with a motion ratio. To sample from the location with the highest intensity of human motions, in line 2, from $\Xi_{\mathrm{near}}$, we select the SWGMM $\xi$ with highest motion ratio.
In line 3 of \cref{alg:sample}, from $\xi$, an SWND is sampled from the selected SWGMM, based on the mixing factor $\pi$. A velocity is drawn randomly from the sampled SWND. Finally, the direction of the sampled velocity is returned and used for motion prediction.


With the direction sampled from the CLiFF-map
, we predict the velocity ($\rho_t$, $\theta_t$) in line 7 of \cref{alg:LHMPAlgo} assuming that a person tends to continue walking with the same speed as in the last time step, $\rho_t = \rho_{t-1}$, and bias the direction of motion with the sampled direction $\theta_s$ as:

\begin{equation}
\begin{gathered}
\theta_t = \theta_{t-1} + (\theta_s - \theta_{t-1}) \cdot K(\theta_{s} - \theta_{t-1}), \\
\end{gathered}
\end{equation}

\noindent where $K(\cdot)$ is a kernel function that defines the degree of impact of the CLiFF-map. We use a Gaussian kernel with a parameter $\beta$ that represents the kernel width:
\begin{equation} \label{eq-kernel}
    K(x) =  e ^ {-\beta \left\Vert x \right\Vert ^ 2}. \\
\end{equation}

\begin{figure}[!t]
\centering
\includegraphics[width=0.9\linewidth]{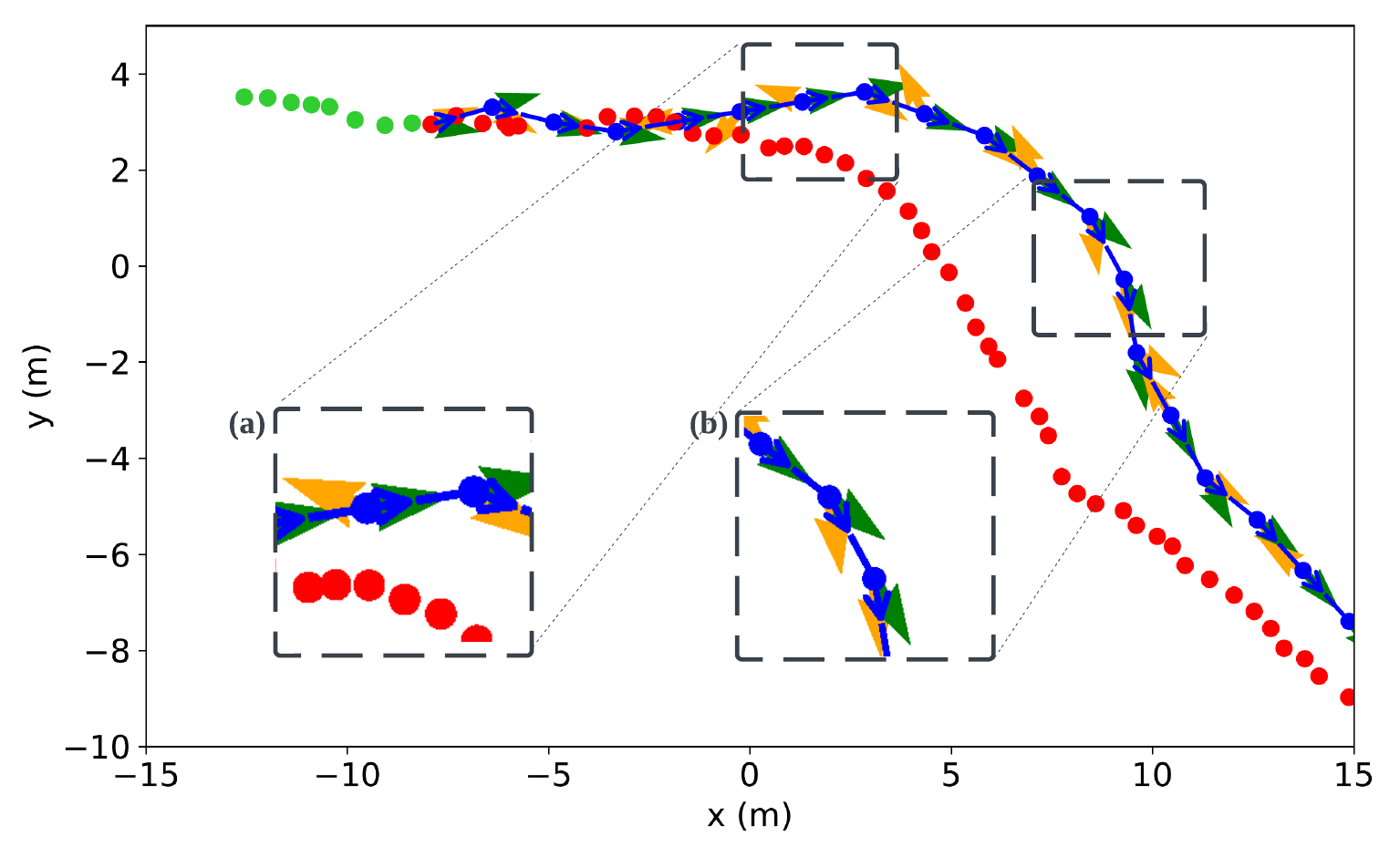}
\caption{Example predictions that visualize the adaptive influence of the CLiFF-map and the constant velocity model on the prediction, based on the sampled direction. \textbf{Green} dots show the observed past states $\mathcal{H}$, \textbf{red} dots show the ground truth future states and \textbf{blue} dots show the predicted states $\mathcal{T}$. In each predicted state, the \textbf{orange} arrow shows the sampled direction from the CLiFF-map $\theta_s$ and the \textbf{green} arrow shows the direction from the last time step $\theta_{t\textendash1}$. \textbf{Blue} arrows between predicted states show the direction of the predicted trajectory. In locations like (\textbf{a}) where the sampled CLiFF-map direction greatly opposes the CVM prediction, the CVM prediction is trusted more. In locations like (\textbf{b}) where the sampled CLiFF-map direction is close to the CVM prediction, the CVM prediction is biased more towards the CLiFF-map direction.}
\label{fig:method-direction1}
\vspace*{-6mm}
\end{figure}

\noindent An example of velocity prediction results is shown in \cref{fig:method-direction1}. With kernel $K$, we scale the CLiFF-map term by the difference between the direction sampled from the CLiFF-map and the current direction according to the CVM. The sampled direction is trusted less if it deviates more from the current direction.
A larger value of $\beta$ makes the proposed method behave more like a CVM, and with a smaller value of $\beta$, the prediction will follow the CLiFF-map more closely.

In the end of each iteration, we add $s_t$ to the predicted trajectory $\mathcal{T}$ (line 9 of \cref{alg:LHMPAlgo}) and update $t$ for the next iteration. After iterating for $T_p$ times, the output is a sequence $\mathcal{T}$ of future states that represents the predicted trajectory.

\section{EXPERIMENTS} \label{section-experiments}
This section describes the experimental setup for qualitative and quantitative evaluation of our CLiFF-LHMP approach.
Accurate map-aware long-term motion predictions are typically addressed with Markov Decision Process (MDP) based methods \cite{ziebart2009planning,karasev2016intent,rehder2017pedestrian,Rudenko2018icra,Rudenko2018iros}. Among them, as the baseline for CLiFF-LHMP, we chose the recent IS-MDP approach~\cite{Rudenko2018iros}. We also compare our method with the constant velocity predictor \cite{scholler2019simpler,rudenko2022atlas}.

We evaluate the predictive performance using the following two real-world datasets:
\begin{enumerate}
    \item \textbf{THÖR} \cite{rudenko2020thor}: This dataset captures human motion in a room with static obstacles. It includes two settings: with one obstacle (denoted as THÖR1, see the top row in \cref{fig:thor_predict_example}) and with three obstacles (denoted as THÖR3, see the bottom row in \cref{fig:thor_predict_example}). The size of the room for data collection is 8.4$\times$18.8 \si{\metre}.
    
    \item \textbf{ATC} \cite{brscic2013person}: This dataset contains trajectories recorded in a shopping mall in Japan. The dataset covers a large indoor environment with total area of around \SI{900}{\metre\squared}. The map of the environment is shown in \cref{fig:predict-example-main}.
\end{enumerate}

THÖR1 and THÖR3 both include four rounds of collected data. We use the first round to build the CLiFF-map and use the remaining three rounds for evaluation. After filtering out short trajectories (shorter than the observation horizon $O_s$) for evaluation, there are in total 247 trajectories in the THÖR1 dataset and 327 trajectories in the THÖR3 dataset. This gives us the train-to-test ratio of about 1 to 3 in both THÖR1 and THÖR3.

The ATC dataset consists of 92 days in total. For building the CLiFF-map, we used the data from the first day (Oct. 24th, 2012). From the remaining 91 days, again after filtering out trajectories shorter than the observation horizon $O_s$, we use 1\,803\,303 trajectories that have continuous motion.



We downsampled both datasets to \SI{2.5}{\Hz}. For observation, we take \SI{3.2}{\second} (the first 8 positions) of the trajectory and use the remaining (up to \SI{50}{\second} or 125 positions) as the prediction ground truth. In the parameter analysis, we also evaluate the effect of setting the observation horizon to different values.

Given the area covered by the ATC dataset ($\sim$\SI{900}{\metre\squared}) and the THÖR dataset ($\sim$\SI{150}{\metre\squared}), the size and number of obstacles in THÖR dataset, and the trajectory lengths available in the datasets, we selected the parameters shown in \cref{tab:scenarios} for our quantitative and qualitative experiments. Because the size of obstacles in the THÖR setting is less than \SI{1}{\metre}, we set the grid resolution to \SI{0.5}{\metre} when building the CLiFF-map from the THÖR dataset, in contrast to \SI{1}{\metre} in the ATC dataset. Also, we set the prediction time step $\Delta t$ to \SI{0.4}{\second} for the cluttered THÖR dataset, in contrast to \SI{1}{\second} for the ATC dataset. In the parameter analysis we evaluate the impact of selecting $\Delta t$ on prediction accuracy.




Sampling radius $r_s$ and kernel $\beta$ are the main parameters in CLiFF-LHMP. The value of $r_s$ is set to a multiple of the CLiFF-map grid resolution. For biasing the current direction with the sampled one, we use the default value of $\beta$ = 1 for both datasets. The impact of both parameters is evaluated in the experiments.
Using the ATC dataset, we specifically evaluate the influence of the three parameters (see \cref{fig:parameter-eva}): observation horizon $O_s \in [1.2, 3.2]$ \si{\second}, sampling radius $r_s \in [1,3]$ \si{\metre}, and kernel parameter $\beta \in [0.5,10]$. We also evaluated the influence of the prediction time step $\Delta t \in [0.4, 1.0]$ \si{\second} using the THÖR dataset (see \cref{fig:thor_delta_t}).

\begin{table}[t]
    \centering
    \begin{tabular}{lll}
     \toprule
        \textbf{Parameter}  & \textbf{ATC} & \textbf{THÖR} \\
        \midrule
        observation horizon $O_s$ & 3.2 \si{\second} & 3.2 \si{\second} \\
        kernel parameter $\beta$ & 1 & 1 \\
        sampling radius $r_s$ & 1 \si{\metre} & 0.5 \si{\metre} \\
        prediction horizon $T_s$ & 1--50 \si{\second} & 0.4--12 \si{\second} \\
        prediction time step $\Delta t$ & 1 \si{\second} & 0.4 \si{\second} \\
        CLiFF-map resolution & 1 \si{\metre} & 0.5 \si{\metre} \\
        kernal parameter $\sigma$ & 1.5 & 1.5 \\
        number of predicted trajectories $k$ & 20 & 20 \\
        \bottomrule
    \end{tabular}
    \caption{Parameters used for evaluation in the ATC and THÖR datasets}
    \label{tab:scenarios}
\vspace*{-8mm}
\end{table}

For the evaluation of the predictive performance we used the following metrics: \emph{Average} and \emph{Final Displacement Errors} (ADE and FDE) and \emph{Top-k ADE/FDE}. ADE describes the error between points on the predicted trajectories and the ground truth at the same time step. FDE describes the error at the last prediction time step. \emph{Top-k ADE/FDE} compute the displacements between the ground truth position and the closest of the $k$ predicted trajectories. For each ground truth trajectory we predict $k$~=~20 trajectories. 


We stop prediction according to Alg.~\ref{alg:LHMPAlgo} when no dynamics data (i.e. SWGMMs) is available within the radius $r_s$ from the sampled location (line 6). If one predicted trajectory stops before $T_s$, it will only be included in the ADE/FDE evaluation up to the last available predicted point. When predicting for each ground truth trajectory, the prediction horizon $T_s$ is either equal to its length or \SI{50}{\second} for longer trajectories.

\section{RESULTS} \label{section-results}



In this section, we present the results obtained in ATC and THÖR with our approach compared to two baselines. The performance evaluation is conducted using both quantitative and qualitative analysis, and we further investigate the approach's performance through a parameter analysis.

\subsection{Quantitative Results}
\begin{figure}
\centering
\includegraphics[width=.50\linewidth]{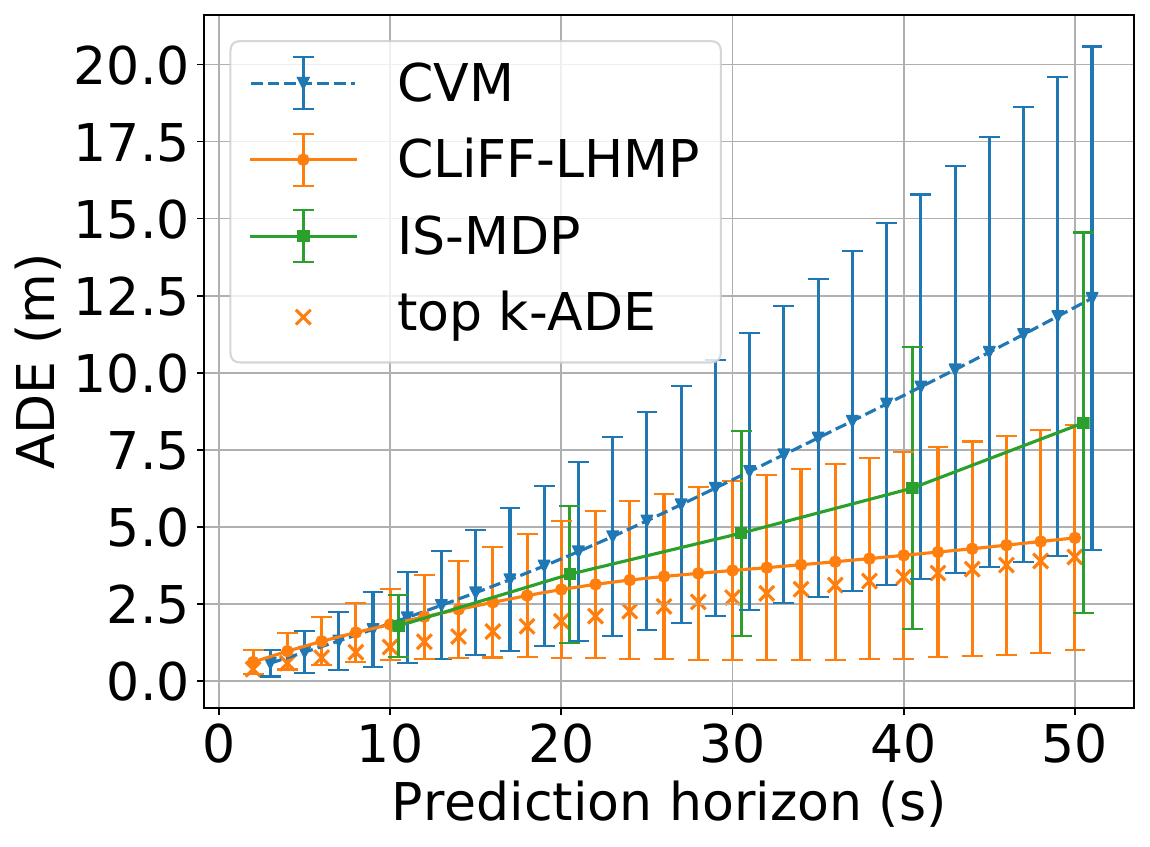}
\includegraphics[width=.48\linewidth]{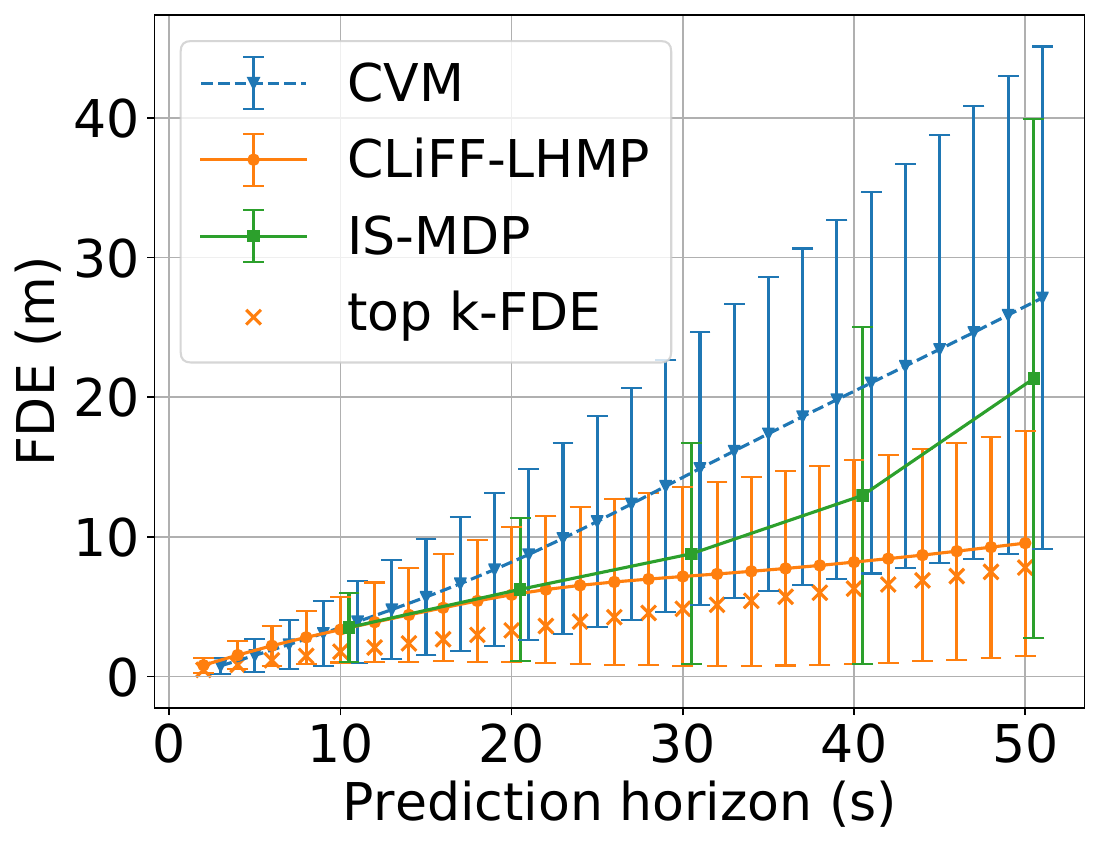}
\caption{ADE/FDE (mean $\pm$ one std. dev.) in the ATC dataset with prediction horizon 1--\SI{50}{\second}.}
\label{fig:atc_final}
\vspace*{-5mm}
\end{figure}

\begin{figure}[t]
\centering
\includegraphics[width=.49\linewidth]{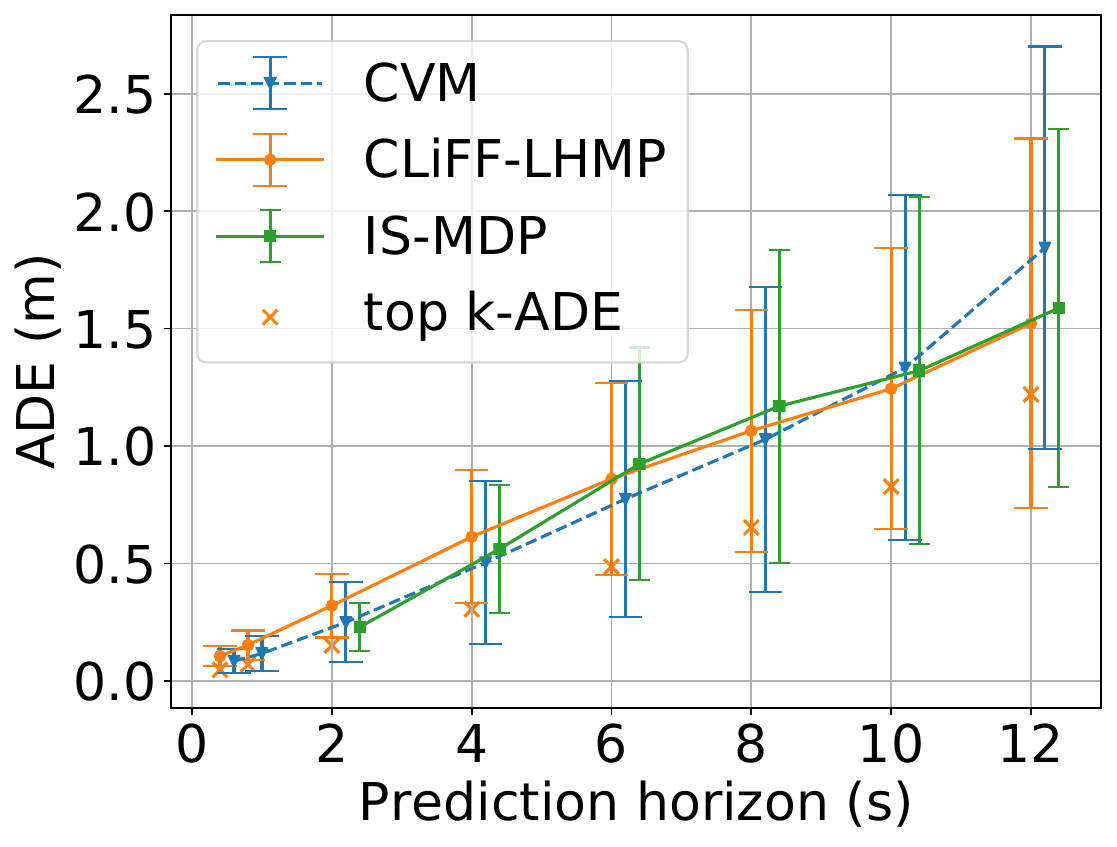}
\includegraphics[width=.47\linewidth]{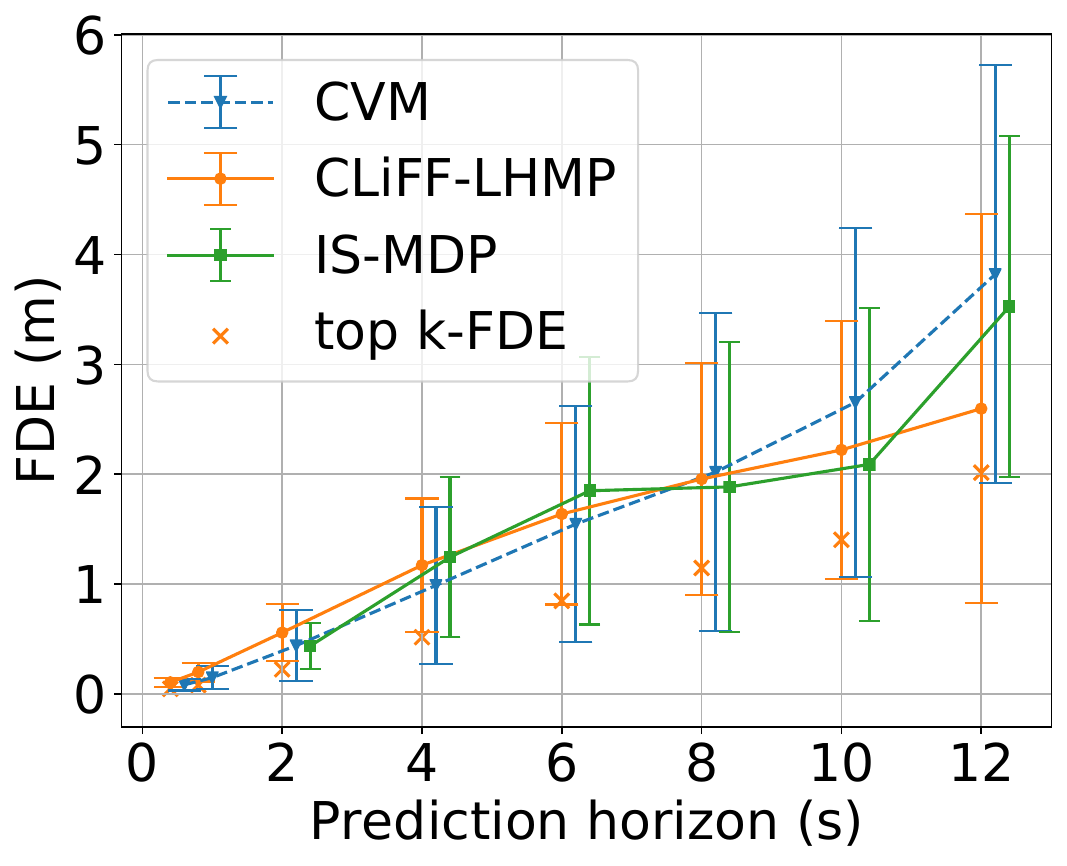}
\includegraphics[width=.49\linewidth]{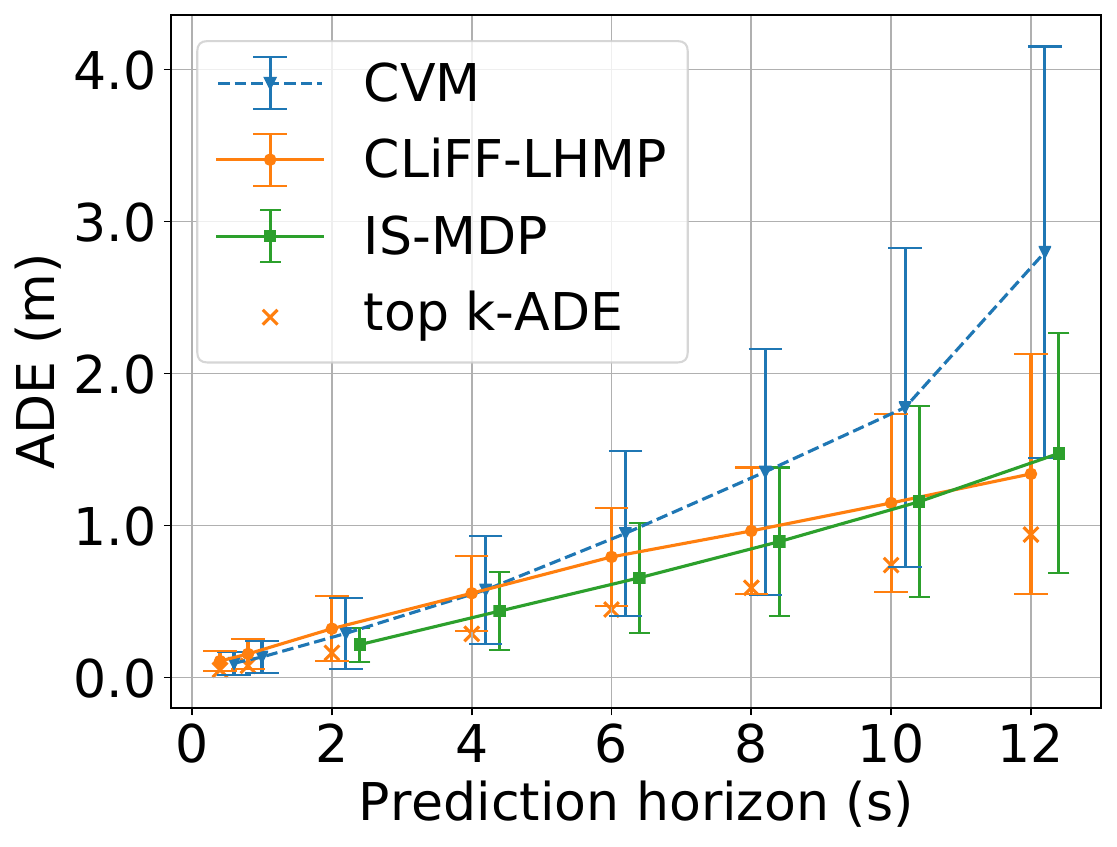}
\includegraphics[width=.475\linewidth]{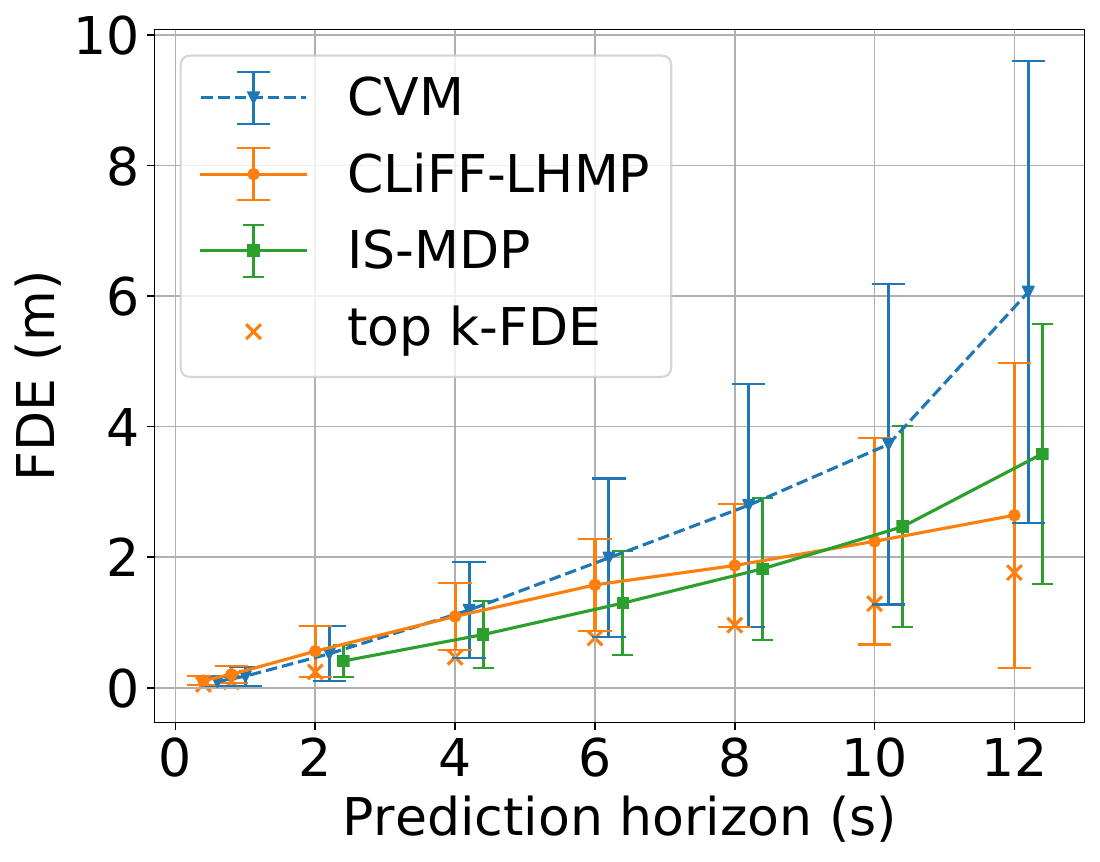}
\caption{ADE/FDE (mean $\pm$ one std. dev.) in the THÖR1 \textbf{(top)} and THÖR3 \textbf{(bottom)} dataset with prediction horizon 0.4--\SI{12}{\second}.}
\label{fig:thor_final}
\vspace*{-5mm}
\end{figure}

\cref{fig:atc_final,fig:thor_final} show the quantitative results obtained in the ATC and THÖR datasets. We compare our CLiFF-LHMP approach with IS-MDP~\cite{Rudenko2018iros} and CVM. In the short-term perspective all approaches perform on par. The mean ADE is marginally lower for CVM compared to the other predictors below \SI{6}{\second} in ATC, below \SI{10}{\second} in THÖR1, and below \SI{4}{\second} in THÖR3. In THÖR3 there are more obstacles that people need to avoid, while THÖR1 and ATC include more open spaces. In open spaces without obstacles, a constant velocity prediction is often a very good short-term predictor~\cite{rudenko2022atlas}. For our approach which accounts for possible deviations from straight trajectories the ADE for short-term predictions is slightly higher.
For prediction horizons less than \SI{10}{\second}, IS-MDP performs better than CLiFF-LHMP. However, the IS-MDP method requires additional input (goal points and the obstacle map) and its performance strongly depends on both. In contrast, our approach makes predictions without explicit knowledge about goals and implicitly accounts for the obstacle layout, as well as the specific ways people navigate in the environment.

In long-term predictions above \SI{10}{\second}, both CLiFF-LHMP and IS-MDP outperform the CVM method. Our approach is substantially better than IS-MDP when the prediction horizon is above \SI{20}{\second} since it implicitly exploits location-specific motion patterns, thus overcoming a known limitation of MDP-based methods~\cite{Rudenko2018iros}. \cref{tab:expres} summarises the performance results of our method against the baseline approaches at the maximum prediction horizon. Our CLiFF-LHMP approach accurately predicts human motion up to \SI{50}{\second} with a mean ADE of \SI{5}{\metre}. At \SI{50}{\second} in the ATC dataset, our method achieves a 45\% ADE and 55\% FDE improvement in performance compared to IS-MDP. At \SI{12}{\second} in THÖR1 and THÖR3, our method achieves an improvement of 6.3\% and 13.3\% ADE (25.7\%, 27.8\% FDE) over IS-MDP, respectively.

\cref{fig:atc_final,fig:thor_final} also show that the standard deviation of ADE and FDE is generally lower for CLiFF-LHMP predictions, compared to CVM and IS-MDP. This indicates that our approach makes more consistent predictions, both in the short- and long-term perspective.

\begin{table}[t]
    \centering
    \begin{tabular}{lllll}
     \toprule
        \textbf{Dataset} & \textbf{Horizon} & \multicolumn{3}{c}{\textbf{ADE / FDE} (m)} \\
          & & CLiFF-LHMP & IS-MDP & CVM \\
        \midrule
        ATC & 50 \si{\second} & \textbf{4.6} / \textbf{9.6} & 8.4 / 21.3 & 12.4 / 27.1 \\
        THÖR1 & 12 \si{\second} & \textbf{1.5} / \textbf{2.6} & 1.6 / 3.5 & 1.8 / 3.8 \\
        THÖR3 & 12 \si{\second} & \textbf{1.3} / \textbf{2.6} & 1.5 / 3.6 & 2.8 / 6.1 \\
        \bottomrule
    \end{tabular}
    \caption{Long-term prediction horizon results on different datasets. With $O_s = \SI{3.2}{\second}$, error reported are ADE/FDE in meters.}
    \label{tab:expres}
\vspace*{-5mm}
\end{table}

\begin{figure*}[t]
\centering
\includegraphics[width=.3235\linewidth]{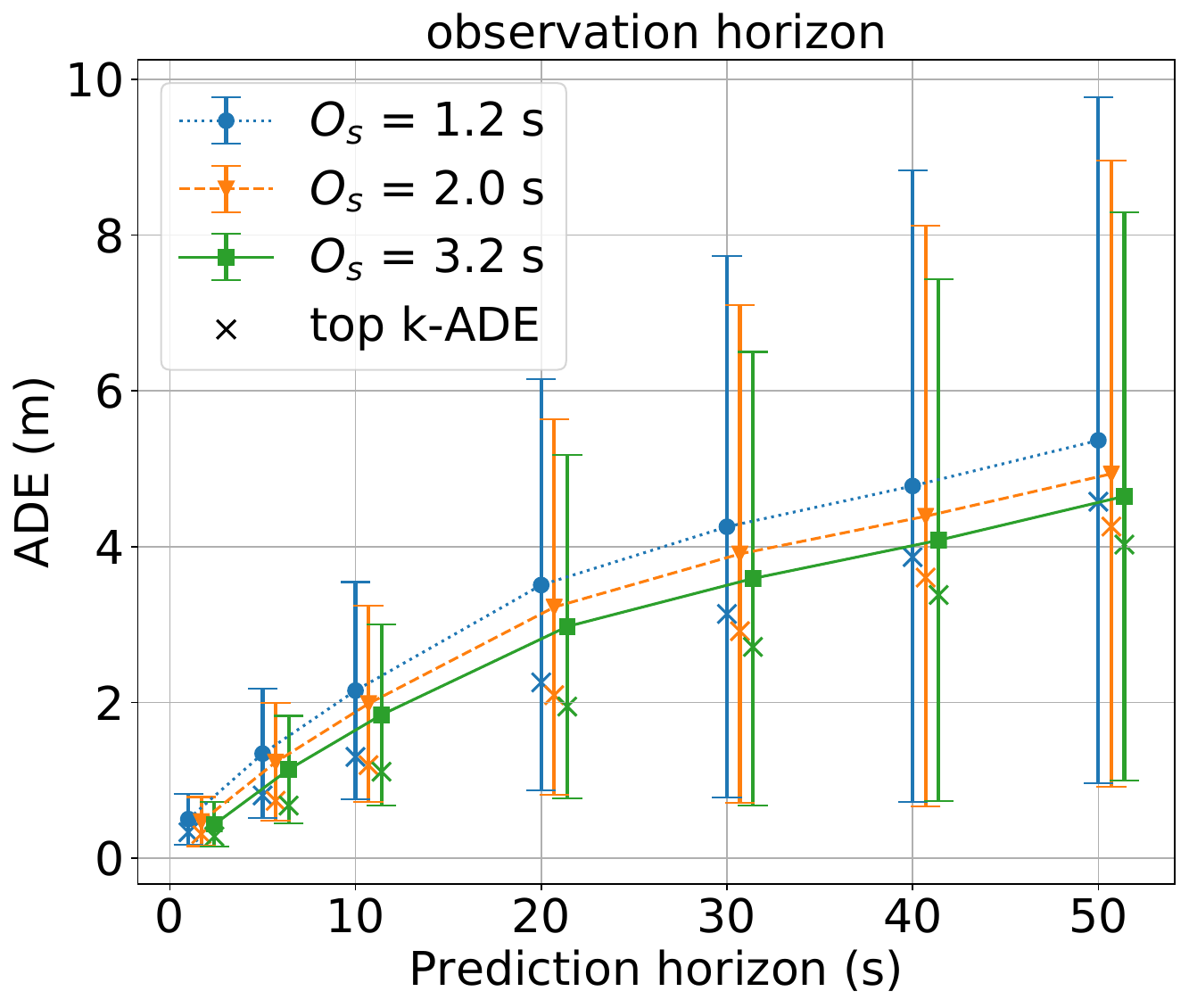}
\includegraphics[width=.3235\linewidth]{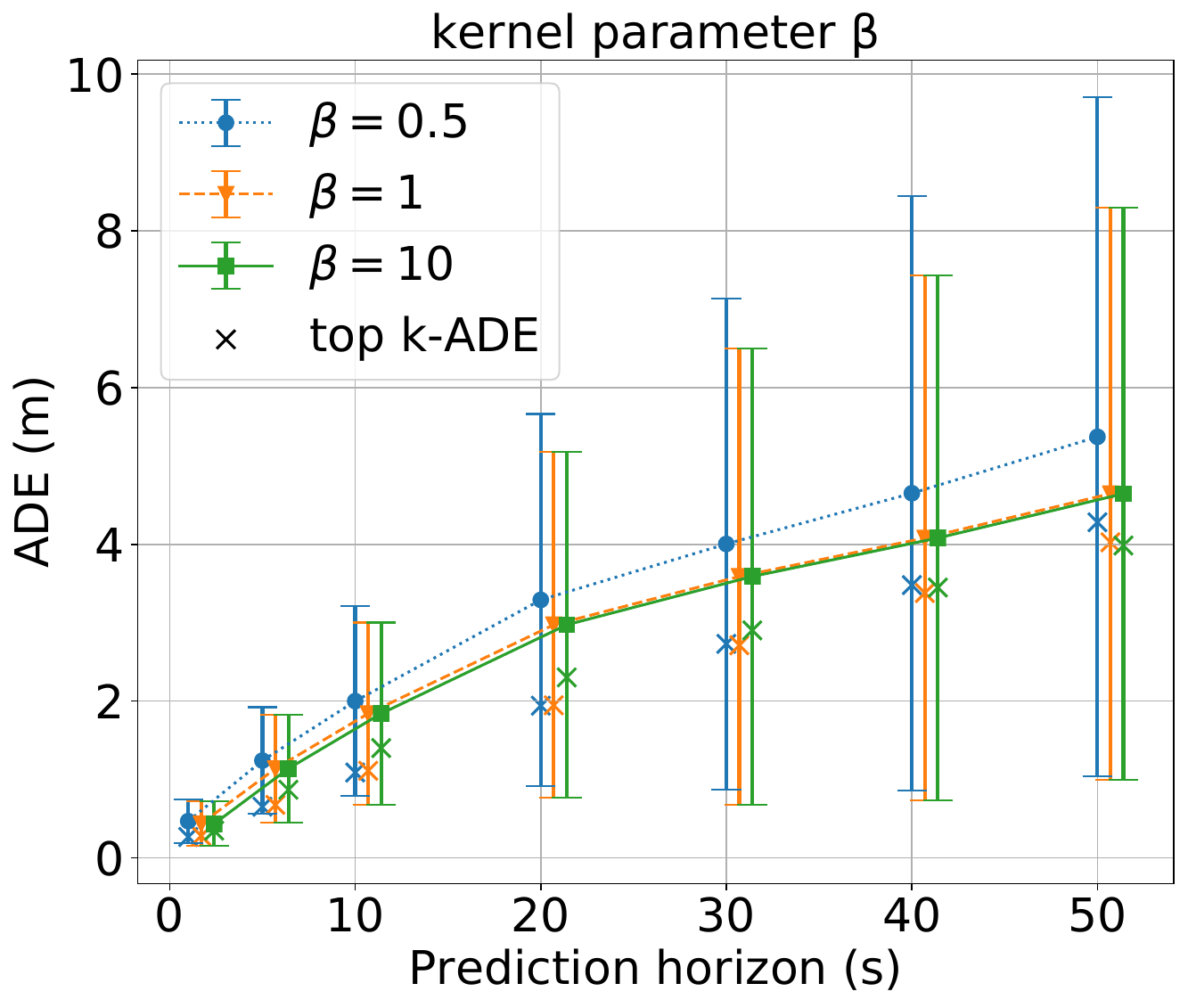}
\includegraphics[width=.315\linewidth]{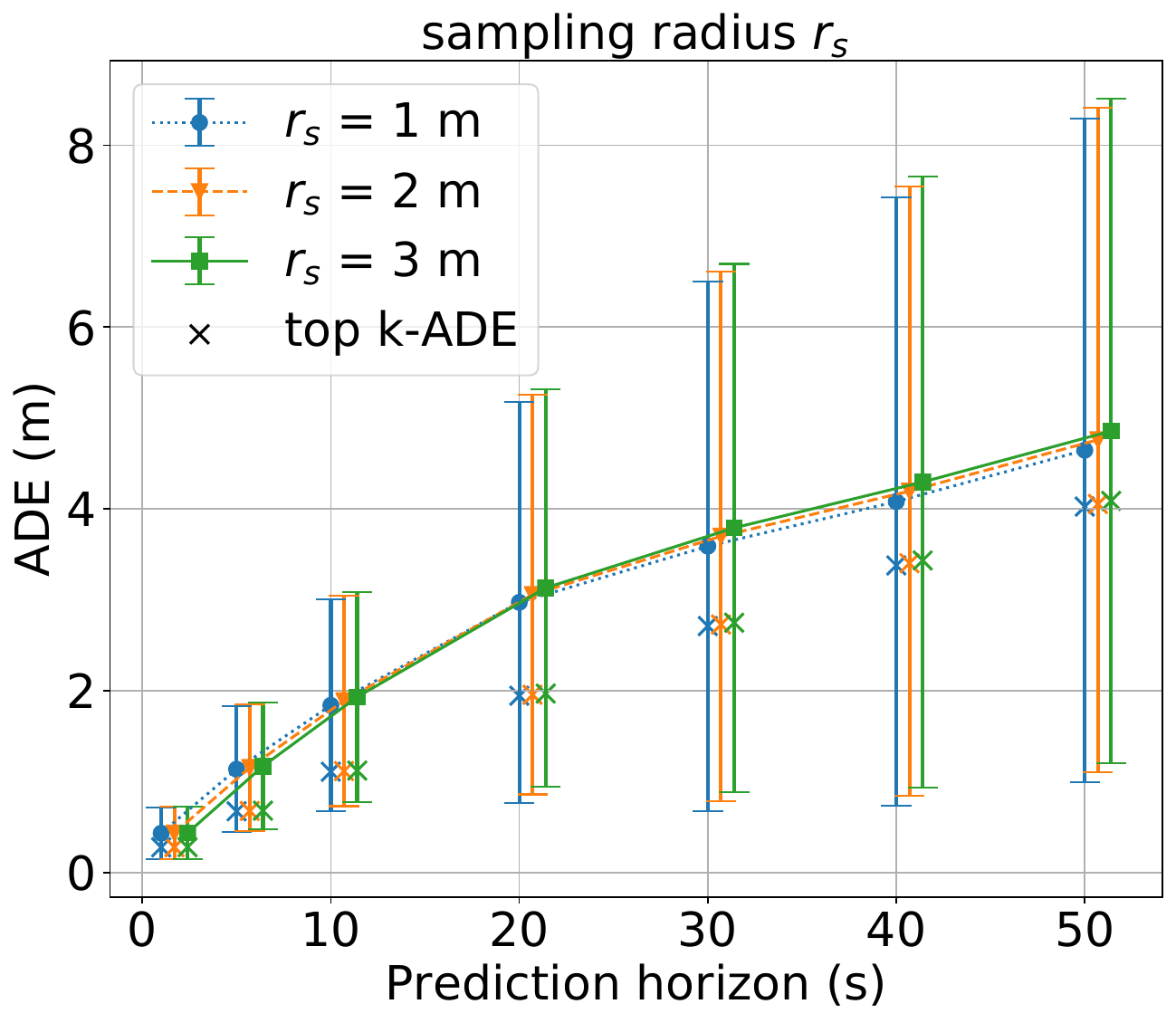}
\caption{
Parameter analysis on the ATC dataset, showing the ADE (mean $\pm$ one std. dev.) over different prediction horizons vs the 
observation horizon $O_s$ \textbf{(left)}, kernel parameter $\beta$ \textbf{(middle)} and sampling radius $r_s$ \textbf{(right)}.}
\label{fig:parameter-eva}
\vspace*{-5mm}
\end{figure*}

\subsection{Parameter Analysis}
In the experiments with different observation horizons (see \cref{fig:parameter-eva}, left), our method performs robustly when the observation horizon is as low as \SI{1.2}{\second}. In the experiments with different $\beta$ values (see \cref{fig:parameter-eva}, middle), we find that $\beta = 1$ is a good trade-off. Lower $\beta$ values make the predictor trust the CLiFF-map more, which can lead to jumps between distinct motion patterns. Setting $\beta$ to a high value such as 10 slightly improves the performance in short-term predictions, however,
as for the CVM model, the CLiFF-LHMP predictor with high values of $\beta$ is prone to fail delivering long-term predictions.
The reason is that we stop predicting when the CLiFF-map is not any longer available close to the predicted location. So, if more trust is put on the CVM component, many ground truth trajectories cannot be predicted successfully for long prediction times. When the planning horizon is set to \SI{50}{\second}, 84\% of ground truth trajectories can be predicted successfully with $\beta = 1$, while with $\beta = 10$, the ratio drops to 52.3\%. Also when the prediction is dominated by the CVM component, the top k-ADE/FDE scores are worse due to a reduced diversity of the predictions.

In the experiments with different values of the sampling radius $r_s$ (see \cref{fig:parameter-eva}, right), we observed a stable prediction performance. Therefore, it is reasonable to set $r_s = 1$ in order to reduce the computation cost. 

\begin{figure}
\centering
\includegraphics[width=.49\linewidth]{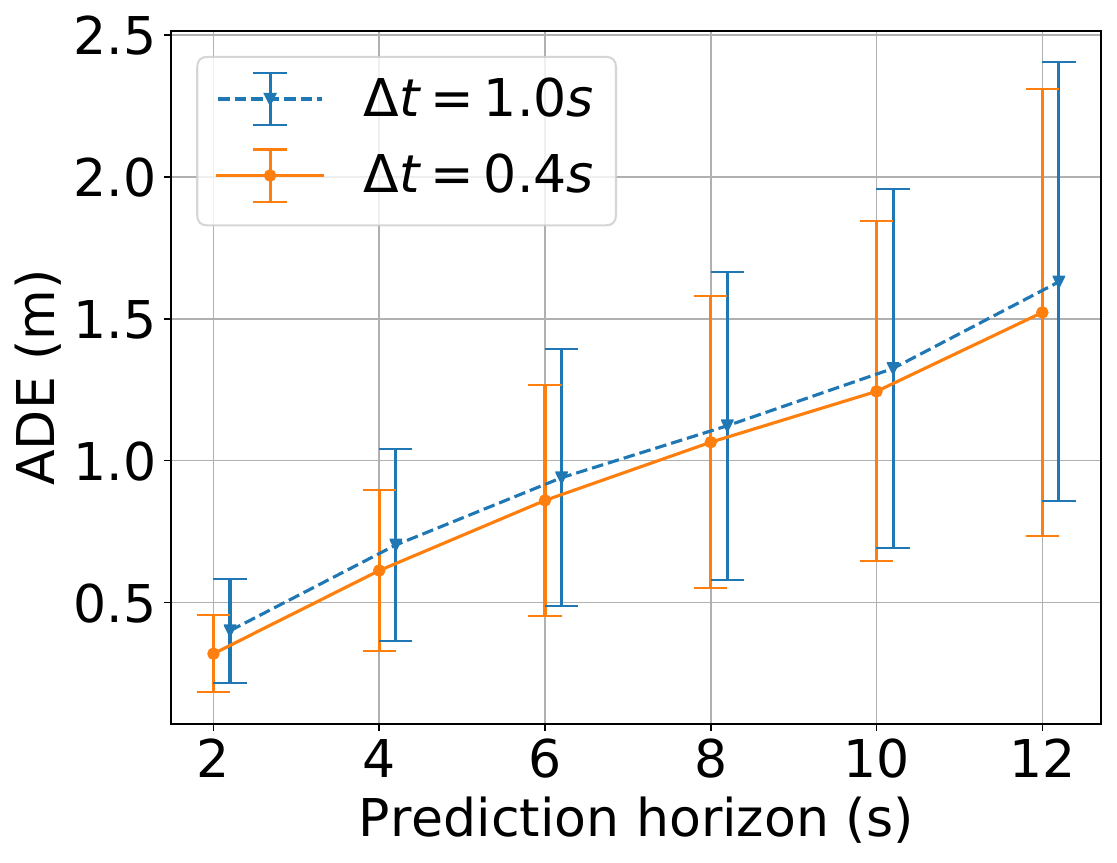}
\includegraphics[width=.49\linewidth]{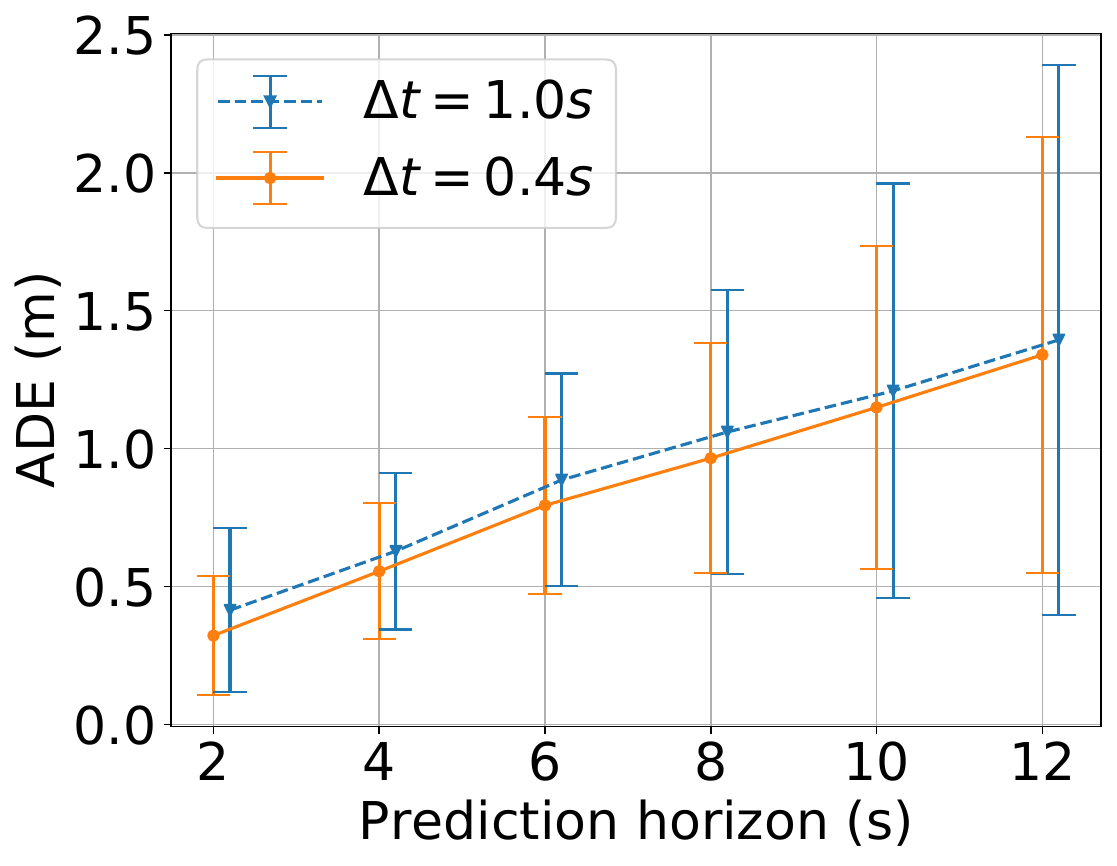}

\caption{Prediction time step $\Delta t$ analysis on THÖR1 \textbf{(left)} and THÖR3 \textbf{(right)} datasets, showing the ADE (mean $\pm$ one std. dev.) over different prediction horizons.}
\label{fig:thor_delta_t}
\vspace*{-3mm}
\end{figure}

In our experiments with the prediction time step $\Delta t$, we observe robust performance with slight improvement when making higher frequency predictions ($\Delta t=$\SI{0.4}{\second} vs. \SI{1.0}{\second}, see \cref{fig:thor_delta_t}). Smaller $\Delta t$ is recommended in cluttered environments, such as in the THÖR dataset. Making iterative predictions with a smaller time step naturally comes at the expense of computational cost increasing linearly for CLiFF-LHMP.
Selecting a larger prediction time step $\Delta t=$\SI{1.0}{\second} drops the performance in THÖR by only approx. 5\% at the maximum prediction horizon, as compared to $\Delta t~=~$\SI{0.4}{\second}.



\subsection{Qualitative Results}
\begin{figure*}
\centering


\includegraphics[clip,trim=0mm 7mm 0mm 0mm,height=40mm]{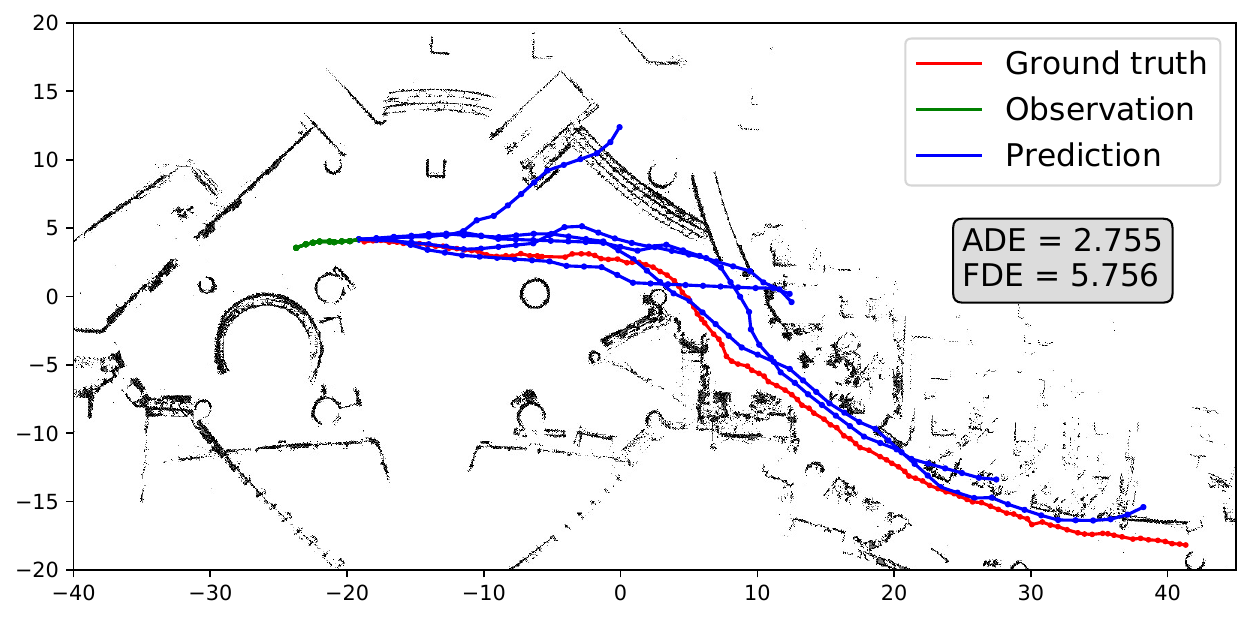}%
\includegraphics[clip,trim=10mm 7mm 0mm 0mm,height=40mm]{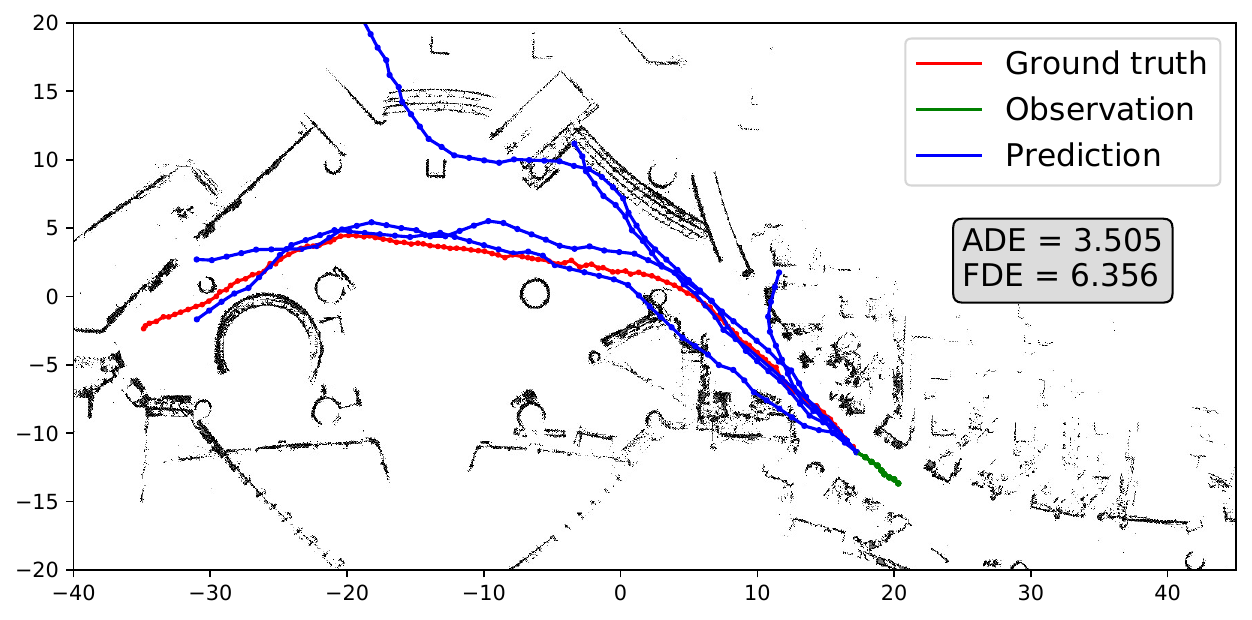}\\
\includegraphics[clip,trim=0mm  0mm 0mm 0mm,height=42.9mm]{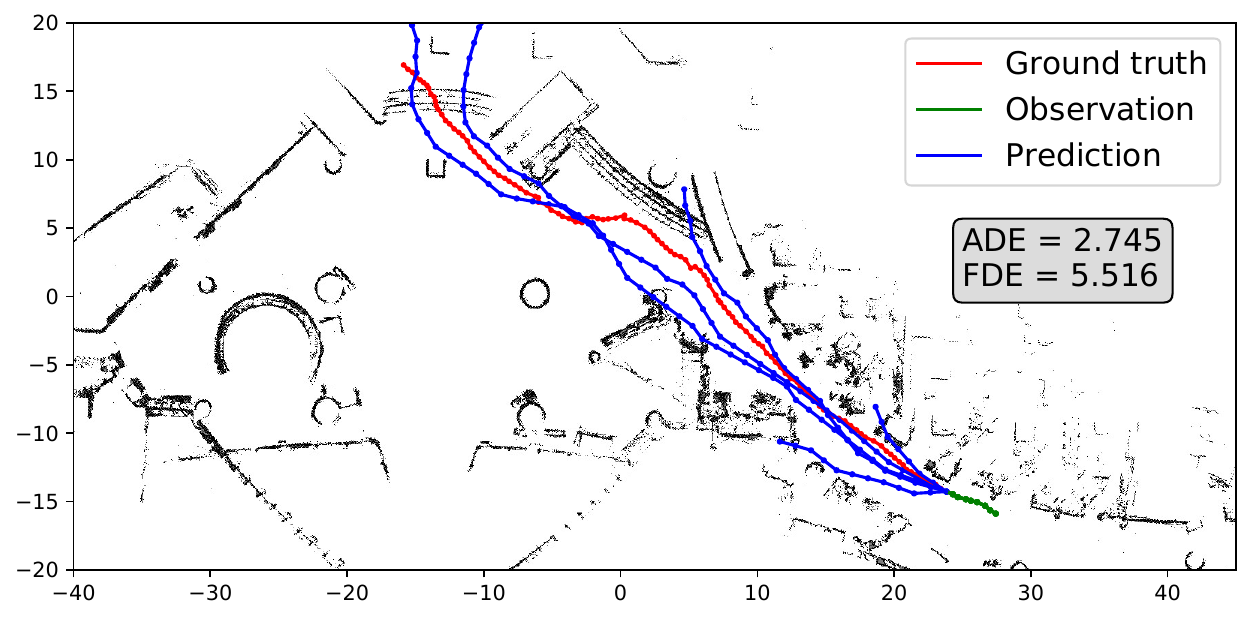}%
\includegraphics[clip,trim=10mm 0mm 0mm 0mm,height=42.9mm]{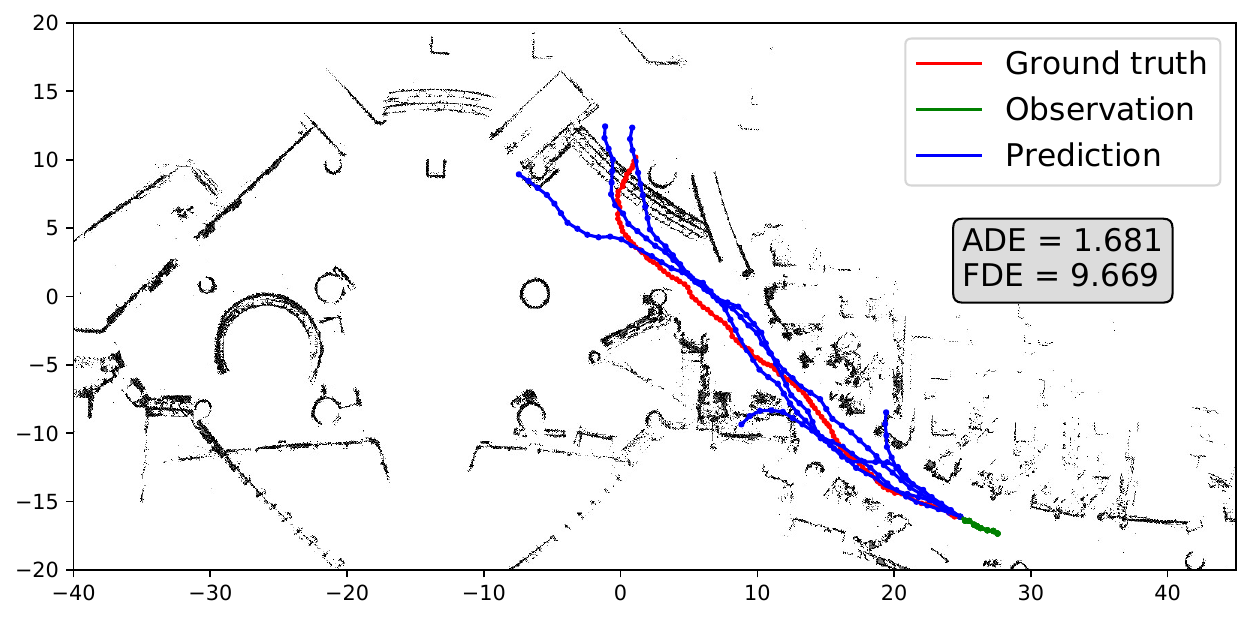}

\caption{Predictions in ATC with $T_s=50$ \SI{}{\second}. \textbf{Red} line shows the ground truth trajectory. \textbf{Green} line shows the observed trajectory and \textbf{blue} lines show the predicted trajectories.
  Note that 
  we  correctly predict trajectories crossing obstacles such as stairs (top of the map) and exits (left of the map).
}
\label{fig:atc_predict_example}
\vspace*{-5mm}
\end{figure*}

\begin{figure}
\centering

\includegraphics[clip,trim=  0mm 17.5mm 0mm 0mm,height=35.5mm]{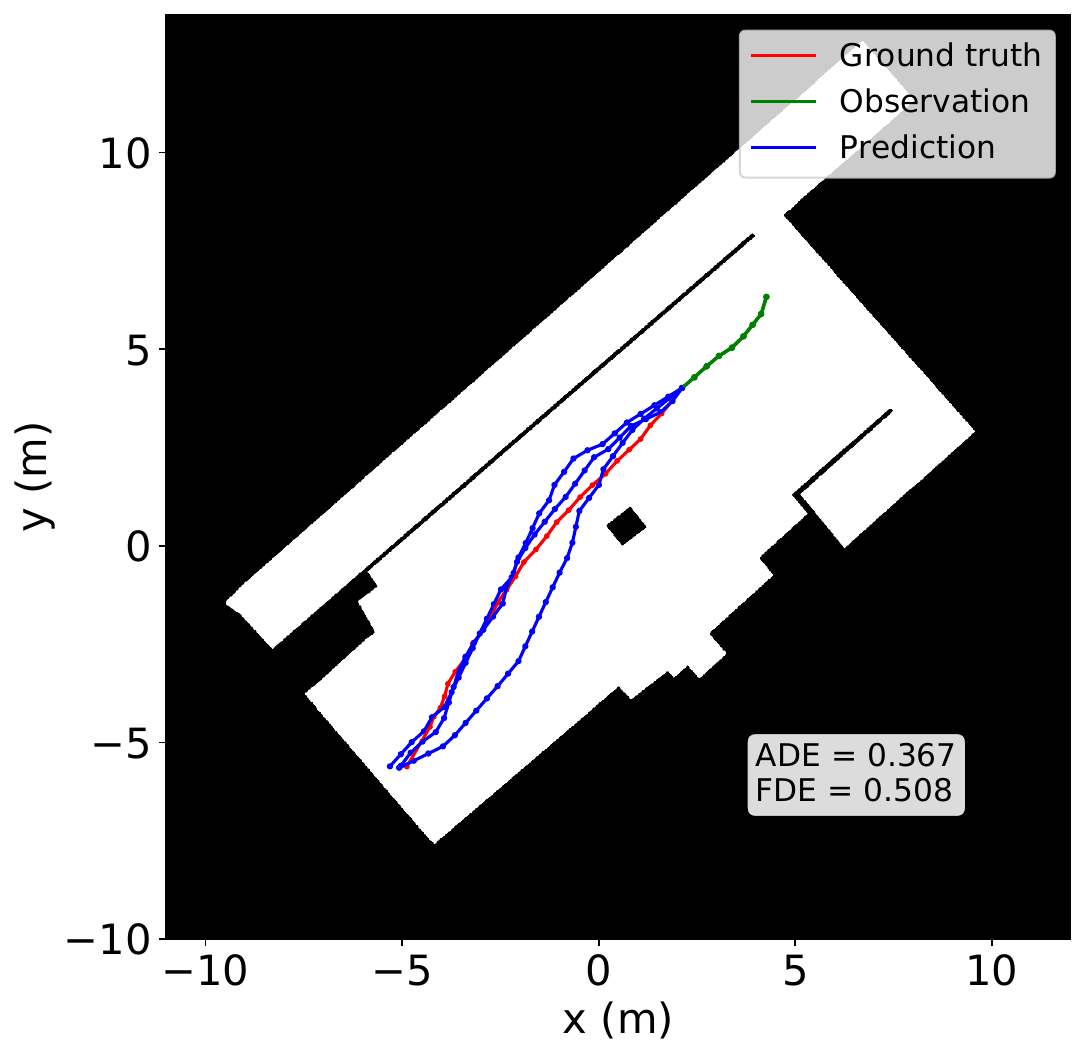}%
\includegraphics[clip,trim= 25.5mm 17.5mm 0mm 0mm,height=35.5mm]{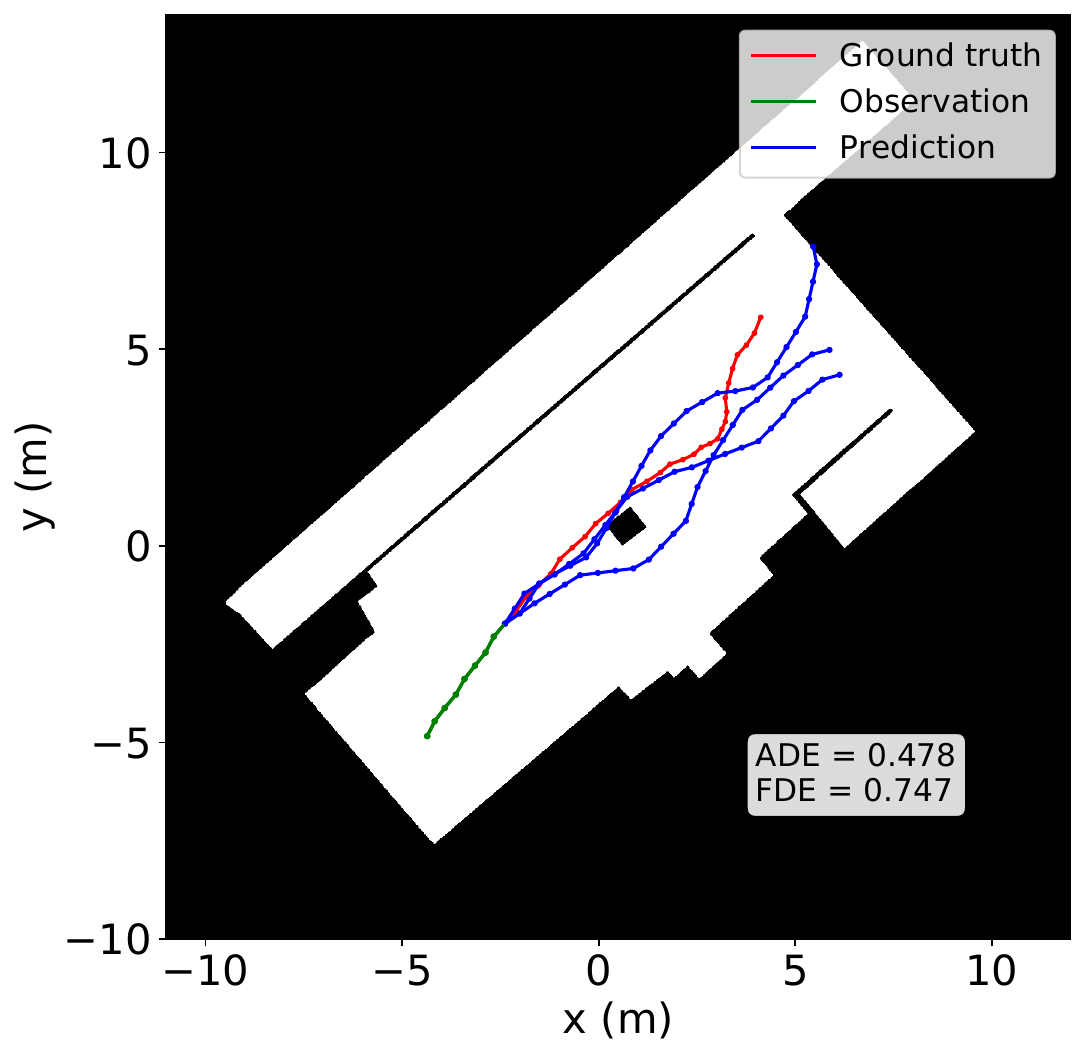}\\
\includegraphics[clip,trim=  0mm  0mm 0mm 0mm,height=39.5mm]{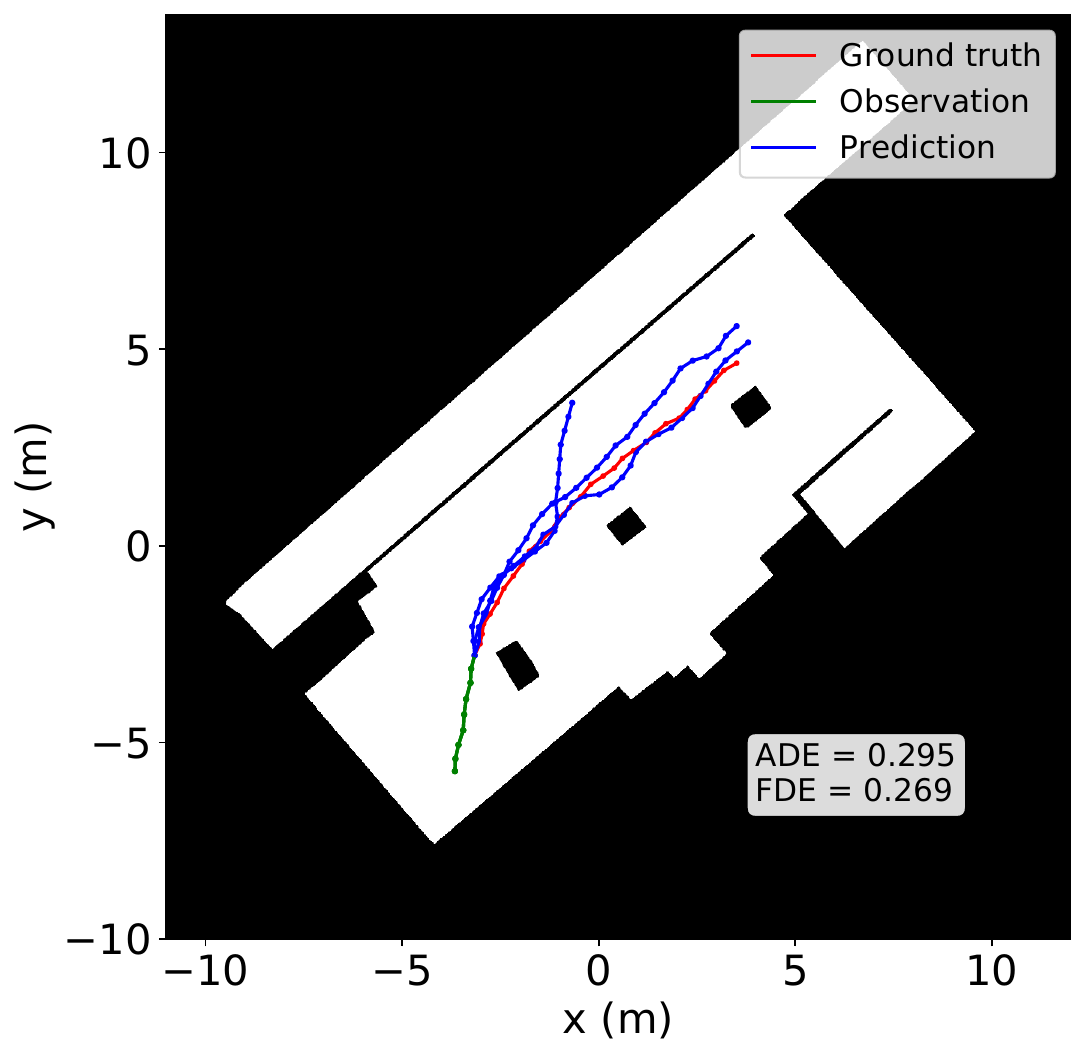}%
\includegraphics[clip,trim= 25.5mm  0mm 0mm 0mm,height=39.5mm]{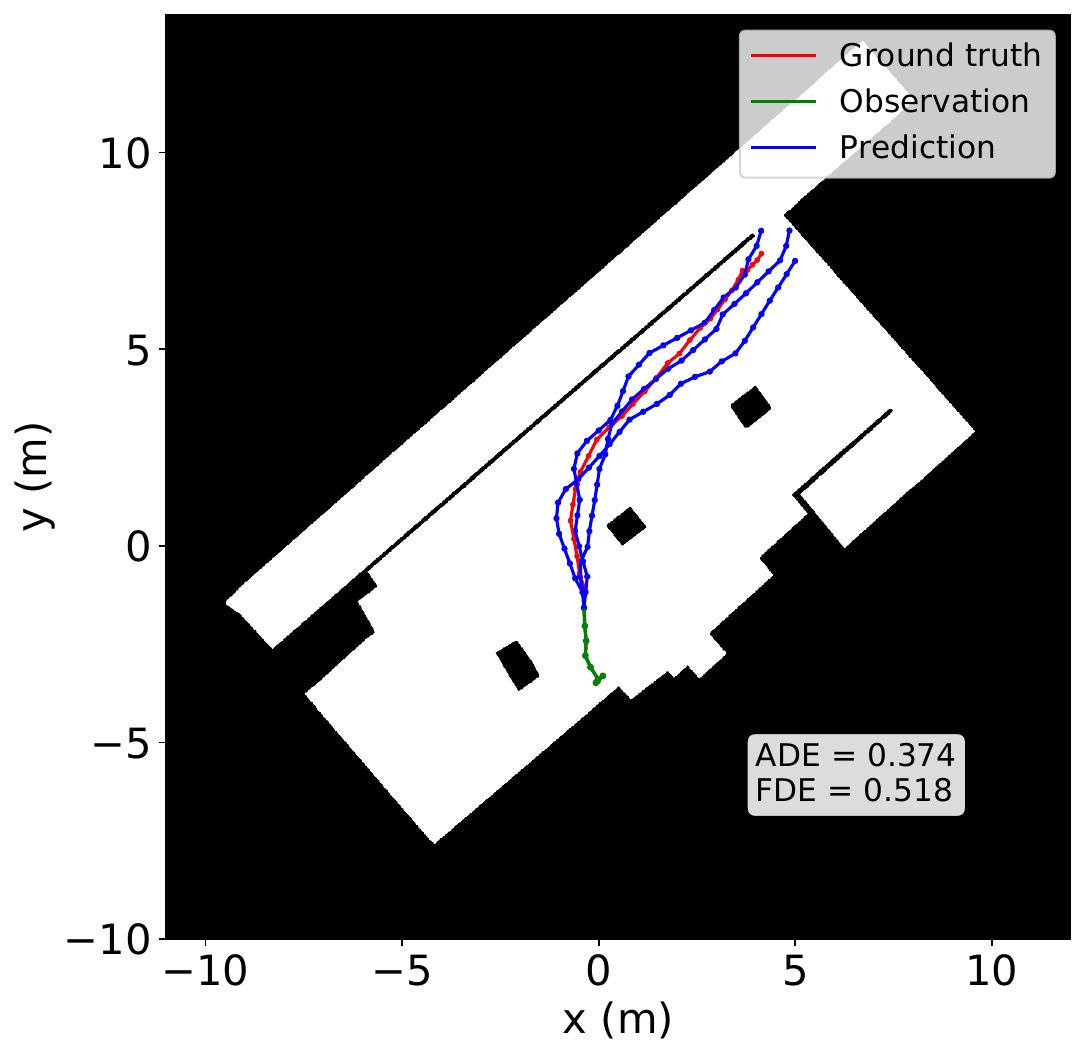}
\caption{Predictions in THÖR1 \textbf{(top)} and THÖR3 \textbf{(bottom)} with $T_s=12$~s. \textbf{Red} line shows the ground truth trajectory. \textbf{Green} line shows the observed trajectory and \textbf{blue} lines show the predicted future trajectories}
\label{fig:thor_predict_example}
\vspace*{-5mm}
\end{figure}

\Cref{fig:atc_predict_example,fig:thor_predict_example} show qualitative results with example predictions. Our approach correctly captures the motion patterns in each scenario, utilizing the environment information during the prediction. \Cref{fig:thor_predict_example} shows that the predicted trajectories avoid the obstacles, even though an obstacle map is not used for predictions. Furthermore, using maps of dynamics built from the observations of human motion makes it possible to predict motion through regions which appear as obstacles in an occupancy map, for example across stairs and through narrow passages (see~\cref{fig:atc_predict_example}). Similarly, using the MoD input keeps predictions in more intensively used areas of the environment, avoiding semantically-insignificant and empty regions, e.g., corners of the room (see~\cref{fig:thor_predict_example}).


\section{CONCLUSIONS} \label{section-conclusions}
In this paper we present the idea to use \emph{Maps of Dynamics} (MoDs) for long-term human motion prediction. By using MoDs, motion prediction can utilize previously observed spatial motion patterns that encode important information about spatial motion patterns in a given environment. We present the CLiFF-LHMP approach to predict long-term motion using a CLiFF-map -- a probabilistic representation of a velocity field from isolated and possibly sparse flow information (i.e. complete trajectories are not required as input). In our approach, we sample directional information from a CLiFF-map to bias a constant velocity prediction.


We evaluate CLiFF-LHMP with two publicly available real-world datasets, comparing it to several baseline approaches.
The results demonstrate that our approach can predict human motion in complex environments over very long time horizons. 
Our approach performs on-par with the state of the art for shorter periods (\SI{10}{\second}) and significantly outperforms it in terms of ADE and FDE for longer periods of up to \SI{50}{\second}. 
We also showed that our method makes more consistent predictions and is not strongly sensitive to the observation horizon.
By exploiting the learned motion patterns encoded in the CLiFF MoD, our method can implicitly infer common goal points and correctly predict trajectories that follow the complex topology of the environment, e.g., navigating around corners or obstacles, or passing through narrow passages such as doors.



Future work will include experimenting with other types of MoDs and motion prediction methods, sampling speed in addition to direction from the MoD, extending CLiFF-LHMP to multi-agent prediction, extending the evaluation to outdoor datasets, as well as estimating confidence values for the predicted trajectories.

\printbibliography

\end{document}